\documentclass[journal,twocolumn,twoside]{IEEEtran}    

\usepackage{hyperref}

\ifCLASSINFOpdf
  \usepackage[pdftex]{graphicx}
  
\else
 \fi

\usepackage{float}
\usepackage{amsmath}
\usepackage{multirow}
\usepackage{color}

\usepackage{graphicx}

\newcommand{\preprintcomment}[2]{\textcolor{blue}{#2}}

\begin{document}

\title{{OpenStreetMap: Challenges and Opportunities in Machine Learning and Remote Sensing}}

\author{John~E.~Vargas-Mu{\~{n}}oz$^*$,
	Shivangi Srivastava$^*$,
	Devis Tuia, ~\IEEEmembership{Member,~IEEE},\\
        and~Alexandre~X.~Falc{\~{a}}o, ~\IEEEmembership{Member,~IEEE}
\thanks{John E. Vargas Mu{\~{n}}oz, and Alexandre X. Falc{\~{a}}o are with the Laboratory of Image Data Science, Institute of Computing, University of Campinas, Brazil.
E-mail: john.vargas@ic.unicamp.br (corresponding author)}
\thanks{Shivangi Srivastava and Devis Tuia are with the Laboratory of Geo-information Science and Remote Sensing, Wageningen University \& Research, the Netherlands.}
\thanks{$^*$These authors contributed equally to this manuscript.}
}

\markboth{IEEE GEOSCIENCE AND REMOTE SENSING MAGAZINE - PREPRINT. DOI: \url{https://doi.org/10.1109/MGRS.2020.2994107}}{Vargas et al. - PREPRINT. DOI: \url{https://doi.org/10.1109/MGRS.2020.2994107}}%
{}

\maketitle

\begin{abstract}
\preprintcomment{}{This is the preprint version. To read
the final version please go to IEEE Geoscience and Remote
Sensing Magazine on IEEE XPlore: \url{https://doi.org/10.1109/MGRS.2020.2994107}}.
OpenStreetMap (OSM) is a community-based, freely available, editable map service that was created as an alternative to authoritative ones. Given that it is edited mainly by volunteers with different mapping skills, the completeness and quality of its annotations are heterogeneous across different geographical locations. Despite that, OSM has been widely used in several applications in {Geosciences}, Earth Observation and environmental sciences. In this work, we present a review of recent methods based on machine learning to improve and use OSM data. Such methods aim either 1) at improving the coverage and quality of OSM layers, typically using GIS and remote sensing technologies, or 2) at using the existing OSM layers to train models based on image data to serve applications like navigation or {land use} classification. We believe that OSM (as well as other sources of open land maps) can change the way we interpret remote sensing data and that the synergy with machine learning can scale  participatory map making and its quality to the level needed to serve global and up-to-date land mapping. A preliminary version of this manuscript has been presented in the first author’s dissertation \cite{phd_john}.
\end{abstract}


\IEEEpeerreviewmaketitle

\section{Introduction}
\label{sec:introduction}

\IEEEPARstart{M}{apping} systems need to be reliable and frequently updated, which makes them costly to be maintained. Due to limited budget, authoritative maps are usually not fully updated at regular time intervals, and present temporal, spatial, and completeness inaccuracies. Recently, Volunteered Geographic Information (VGI)~\cite{Goodchild_2007} has appeared as an alternative to authoritative map services. VGI collects mapping information from individuals, usually volunteers, and stores the information in a database which is often freely available. OpenStreetMap (OSM) is one of the most successful VGI projects. It started in 2004 and currently counts more than 5 million users~\footnote{https://osmstats.neis-one.org/} from different parts of the world. This gives OSM the potential to provide updated mapping data at global scale. 

OSM information is represented by four types of data: nodes, ways, relations, and tags, which are constantly edited by volunteers. 
\begin{itemize}
\item[-]A node is a location on the earth's surface, as determined by latitude and longitude coordinates. 
Points of Interests (POIs) such as bus stations can be represented by nodes. 
\item[-]A way is a list of nodes forming polylines, that can represent road networks or areal objects (closed ways) like buildings. 
\item[-]A relation represents the relationship among objects --- e.g., a group of road segments can represent a bus route. 
\item[-]A tag is a key-value pair that contains information of an object --- e.g., a restaurant can be represented by a way with a tag ``amenity=restaurant".
\end{itemize}

Although the OSM data is constantly under improvement, the completeness and quality of the annotations in different regions are affected by the number and mapping skills of the volunteers~\cite{Mooney_2012}. 
As reported in~\cite{Haklay_2010}, the spatial coverage of OSM is heterogeneous in different geographical regions --- i.e., urban areas are more regularly updated than rural areas. In road networks, missing roads are reported in~\cite{Funke_2015} and inaccurate road tags are reported in~\cite{Jilani_2014}. The positional accuracy of building footprints in OSM sometimes requires corrections~\cite{Xu_2017}.
Several works in the literature have studied methods to assess the quality of OSM data by quantifying: data completeness~\cite{Koukoletsos_2012}, positional accuracy~\cite{Fan_2014}, semantic tag acccuracy~\cite{Girres_2010}, and topological consistency~\cite{Neis_2011}. Some works focus on meta analysis of OSM, like the analysis of the contributors' activities~\cite{Neis_2014, Arsanjani_2015exploration} and the quality assessment of the OSM data~\cite{Senaratne_2017, Jilani_2019traditional}.

Despite its completeness and quality issues, OSM has been widely used for several applications: e.g., validation of 
{land cover} maps~\cite{Fonte_2015}; 
{land cover}/{land use} classification~\cite{Srivastava_2018, Audebert_2017, Srivastava_2019understanding}; navigation and routing applications like traffic estimation~\cite{Lin_2018transfer} and pedestrian, bicycle, and wheelchair routing~\cite{Schmitz_2008, Neis_2015}; detection of buildings and roads in aerial imagery~\cite{VargasMunoz_2019, Mnih_2012}; 3D city modelling~\cite{Over_2010}; indoor mapping~\cite{Goetz_2012}; and location-based map services~\cite{Schilling_2009}.

In the recent years, the automation of tasks involving OSM data has received increasing attention: on one hand, research aiming at the improvement of the OSM layers has turned to Earth observation and machine learning algorithms as automatic ways to complete footprints in missing areas and verify specific annotations, mostly to ease and decrease the workload of volunteers. On the other hand, the rise of deep learning~\cite{Zhu17} has found in OSM a very valuable source of label information to train large models for image recognition from Earth Observation data, but also brought issues related to the quality, standardization, and completeness of the data used for training the models. In both cases, and in countless applications, the alliance of machine learning, Earth Observation, and OSM layers is proving to be an enabling factor for tackling global challenges in new ways.

To present the potential and opportunities of OSM for geoscience and remote {sensing} research, in this paper we present a review of methods based on machine learning to improve and use OSM for applications in different domains. Note that, we do not intend to cover an exhaustive list of OSM applications, but we focus on the ones that involve machine learning techniques only. 

In Section~\ref{sec:improving_osm_data}, we review methods based on machine learning to improve the completeness and quality of the OSM objects, such as building footprints,  street networks, and points of interest. Section~\ref{sec:osm_applications} reviews the works based on machine learning for applications like 
{land cover}/{land use} classification, navigation, and fine-scale population estimation. Section~\ref{sec:future_works} summarizes the discussion and draws promising future research areas at the interface of ML and VGI. Section~\ref{sec:conclusions} states the conclusion about this paper.


\section{Improving OSM data with machine learning}
\label{sec:improving_osm_data}

The next sections describe methods based on machine learning to improve different types of OSM annotations: building footprints, street networks, semantic tags, and points of interest. 

\subsection{Building footprints}
\label{subsec:building_footprints}

\vspace{5mm}
\noindent{\textbf{Detecting geometric mismatches}}\\

\begin{figure}[!t]
\begin{center}
\begin{tabular}{cc} 
  \includegraphics[width=0.45\columnwidth]{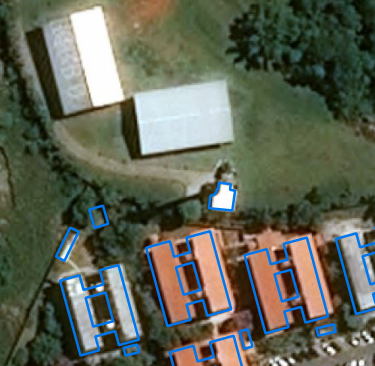} & 
  \includegraphics[width=0.45\columnwidth]{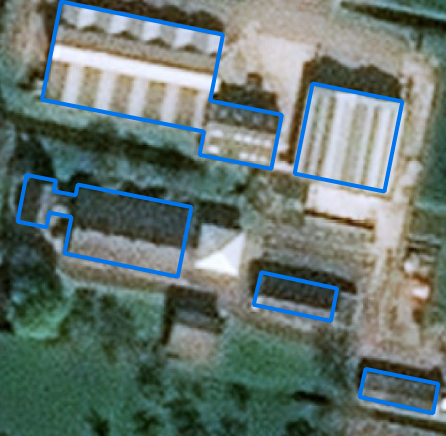} \\
  {a)} & {b)}
\end{tabular} 
\end{center}
\caption{Examples of OSM building annotations (regions highlighted in blue) with completeness errors superimposed over Bing aerial imagery: a) incomplete annotation of buildings in the city of S\~{a}o Paulo, b) incomplete annotation of buildings in the city of Amsterdam.
	\label{fig:incomplete_osm_building_annotations}}
\end{figure}

Buildings are one of the most widely annotated objects in OSM. Although the geometrical features and tags of the buildings in OSM are usually correct (especially in urban areas), there are cases where the building footprints are not accurately mapped by volunteers. Figure~\ref{fig:incomplete_osm_building_annotations} 

presents examples of incomplete OSM building annotations in the cities of S\~{a}o Paulo and Amsterdam. An autoencoder neural network method is proposed in~\cite{Xu_2017} to measure the accuracy of OSM building annotations with respect to official governmental data in the city of Toronto. The authors extract geometrical mismatch features to train an autoencoder neural network. Then, the reconstruction error predicted with the trained model is interpreted as a score that represents the quality of the annotation for a particular region. This method could be useful to import building footprints from other sources to OSM, since the proposed score could be used to identify where are the most mismatched regions that need to be carefully analyzed by annotators.

\vspace{5mm}
\noindent{\textbf{Detecting vandalism behavior}}\\

Some building annotations in OSM are intentionally edited with wrong geometries, such cases are known as digital vandalism. In order to identify vandalism in OSM data, the authors in~\cite{Neis_2012towards} propose a rule-based system that analyzes temporal data of user annotations. For the case of buildings, in particular the authors in~\cite{Truong_2018} propose a clustering-based method to detect vandalism of building annotations. This method extracts geometrical features from the OSM vectorial building data (e.g., perimeter, elongation, convexity, and compacity) and then finds groups in the feaure space to detect outliers, which are assumed to be possible vandalized building footprints.

\vspace{5mm}
\noindent{\textbf{Correct and create new annotations}}\\

\begin{figure}[!t]
\begin{center}
\begin{tabular}{ccc} 
  \includegraphics[width=0.29\columnwidth]{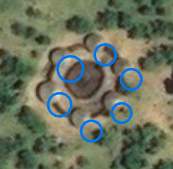} & 
  \includegraphics[width=0.29\columnwidth]{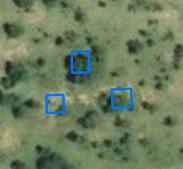} &
  \includegraphics[width=0.29\columnwidth]{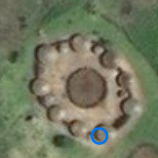} \\
  {a)} & {b)} & {c)}
\end{tabular} 
\end{center}
\caption{Examples of rural building annotations errors found in OSM superimposed over Bing aerial imagery (polygons in blue represent building annotations): a) misalignment annotation errors, b) annotations that do not match with any building, c) missing building annotations.
	\label{fig:osm_building_annotation_error_types}}
\end{figure}

In addition to the geometrical features of OSM building annotations, other works in the literature use aerial imagery to correct building annotations~\cite{VargasMunoz_2019, Zhuo_2018}. In~\cite{VargasMunoz_2019}, the authors propose a methodology to correct rural building annotations in OSM. The paper points out three common problems in OSM building annotations in rural areas: they are geometrically misaligned (see {Fig.}~\ref{fig:osm_building_annotation_error_types}a), some annotations do not correspond to buildings in the updated aerial imagery ({Fig.}~\ref{fig:osm_building_annotation_error_types}b), and some OSM annotations are missing for buildings that are present in the updated aerial imagery ({Fig.}~\ref{fig:osm_building_annotation_error_types}c). The authors propose solutions for the three issues by using  Markov Random Fields (MRF) to align annotations and remove annotations using a building probability map obtained by a Convolutional Neural Network (CNN). The last step of the method is the prediction of new building annotations that are missing by using a CNN model that predicts building footprint with predefined shape priors. The method in~\cite{Zhuo_2018} aims at correcting OSM building annotations by using contour information from image segmentation of oblique images, acquired by Unmanned Aerial Vehicles (UAV). The paper uses contour information of multiple-view images and 3D building models to correct OSM building annotations. Some companies have also made great efforts to improve the
geometrical completeness of OSM. Microsoft has used deep learning models to compute new building footprints by processing satellite imagery in the United States of America~\footnote{https://blogs.bing.com/maps/2018-06/microsoft-releases-125-million-building-footprints-in-the-us-as-open-data}.

\subsection{Street Network}
\label{subsec:street_network}

The quality of street network information in OSM is crucial for several applications. The quality of street type tags and the completeness of the road network data are critical for route planning, while the street names are important to perform queries on the OSM map. As reported in~\cite{Neis_2011, Barrington_2017world}, road networks in OSM present heterogeneous quality and some completeness errors even in urban areas. 

\vspace{5mm}
\noindent{\textbf{Correcting topology}}\\

In~\cite{Funke_2015} the authors propose a method to improve the completeness of road networks. Specifically, they present a method based on machine learning to identify missing roads between candidate locations (two nodes of the OSM road network). The method extracts several features from each pair of candidates from OSM data such as connectivity, street type, and node degree in the OSM road network. The work shows empirically some evidence that the shortest path distance between two nodes in an OSM road network is correlated with the straight line distance. The extracted features are then used to train a Logistic Regression classifier to predict missing roads. The last step involves the pruning of some predictions to increase their precision. 

\vspace{5mm}
\noindent{\textbf{Extracting roads from aerial images}}\\

The automatic extraction of road networks has also been attempted by analyzing remote sensing imagery. The work proposed in~\cite{Cheng_2017automatic} performs per pixel classification using a CNN-based method and later {obtains} the centerline of the roads.
The authors of~\cite{Mattyus_2017deeproadmapper} also use CNNs and centerline computation, but additionally they correct some gaps in the extracted road network by generating several possible missing road candidates and selecting some of them with the help of another CNN. Recently, more accurate results have been obtained by some methods~\cite{Bastani_2018roadtracer,Ventura_2018iterative} that iteratively construct the road network graph, by adding new edges to the graph. The authors in~\cite{Bastani_2018maid} propose a method that is as accurate as~\cite{Bastani_2018roadtracer} but much more efficient. This CNN-based method output road directions for each pixel to create a road network. 

The authors of~\cite{Mattyus_2016hd} propose a CNN-based method that uses aerial imagery as well as ground-based pictures, in the city of Karlsruhe, Germany, to extract road networks and other objects such as parking spots and sidewalks that could be integrated into the OSM database.
Facebook has also implemented deep learning methods to analyze satellite imagery but for the detection of new road networks in developing countries in the OSM map~\footnote{https://wiki.openstreetmap.org/wiki/AI-Assisted\_Road\_Tracing}

\begin{figure}[t]
\centering
\includegraphics[width=0.98\columnwidth]{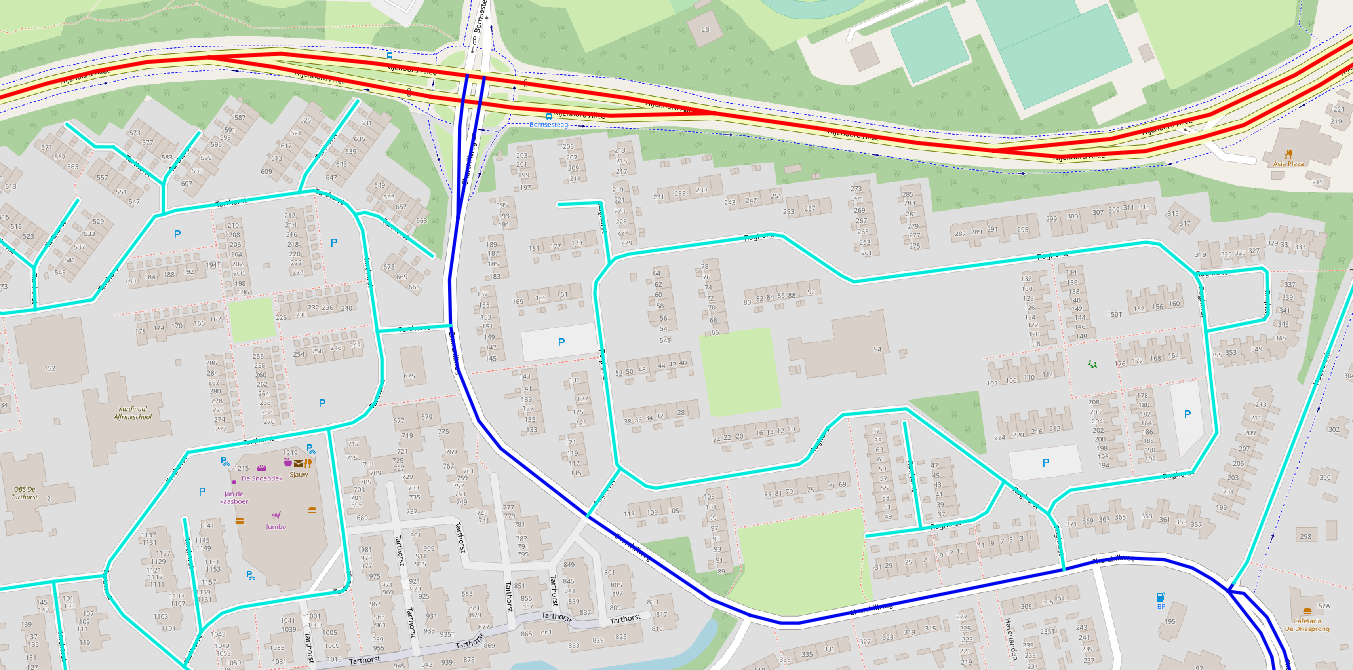}
\caption{Three different street types (secondary, tertiary and residential colored in red, blue, and cyan, respectively) over the OSM map in the city of Wageningen, the Netherlands.\label{fig:street_types}}
\end{figure}

\vspace{5mm}
\noindent{\textbf{Assigning attributes: road types}}\\

Topological and geometrical features extracted from road networks can be very useful to predict street types. Figure~\ref{fig:street_types} depicts three types of streets (i.e., secondary, tertiary, and residential streets) over the OSM map. 
It can be observed that residential street segments are small in length (distinctive geometrical feature) and that tertiary roads are connected with several residential streets (distinctive topological feature).
A solution to fix incorrect street type tags in OSM is then presented in~\cite{Jilani_2013}. The authors extract topological features from OSM road network data to train a neural network classifier that predicts if a street is of type residential or pedestrian. This classifier can be useful to find inconsistent street type tags in OSM.

A multi-granular graph representation of street networks is proposed in~\cite{Jilani_2013multi}. This structure combines the primal (where nodes are road intersections) and dual representations (where nodes are fragments of roads) of road networks. This multi-granular representation is used in~\cite{Jilani_2014} to extract features and train a Random Forest classifier that is able to classify streets to 21 different street categories in OSM. The method uses Bag of Words computed over geometrical and topological features of the analyzed streets and their neighbors. 
The method in~\cite{Jilani_2016probabilistic} uses graphical models with geometrical and spatial features, such that the parameters of the model are learned by Structured Support Vector Machines (SSVM)~\cite{Tsochantaridis_2005largemargin}. More recently, the authors in~\cite{Jilani_2018multi} propose a multi-layer CRF (Conditional Random Field) model to perform hierarchical classification of street types into coarse and fine-grained classes. {In \cite{Jepsen_2019graph} the authors propose a graph convolutional network based method for driving speed estimation of road segments. The output of this method could be used as an additional feature for models that classify road types.}

\vspace{5mm}
\noindent{\textbf{Assigning attributes: structural}}\\ 

Detecting multilane roads is important to model traffic in urban areas. However, the tag 'lanes' that is used to specify how many traffic lanes a road has is usually empty in OSM. Therefore, some works~\cite{Li_2014polygon, Xu_2019multilane} have developed methods to detect multilane roads in OSM data by analyzing the polygons formed by the road network. This is possible because frequently multilane roads are digitized as multiple parallel roads with terminals in road crossings. In~\cite{Li_2014polygon} the authors {propose} to extract geometrical features (e.g., area, perimeter, and compactness) from polygons obtained from the road network and train SVM classifier to predict if a road has multiple lanes. After that a postprocessing step is performed, by using a region growing algorithm, to analyze if roads connected to the predicted multilane roads are also multilane roads. The method proposed in~\cite{Xu_2019multilane} also uses geometrical features to train a classifier, a Random Forest in this case, but the predictions are used to train a second Random Forest classifier that uses geometrical and topological features, such as the percentage of neighboring roads that {are} classified as a multilane road by the first classifier. In~\cite{Mattyus_2015enhancing} the authors use an MRF and data extracted from remote sensing images (e.g., edge information, cars detected, and contextual information) to correct OSM road centerline locations and estimate the width of OSM roads. 

\vspace{5mm}
\noindent{\textbf{Extracting road data from GPS locations}}\\

Sequences of GPS positions (also called tracking data) of users can be used to enrich OSM data. This information can be obtained by GPS locations of cars or applications installed in the volunteers' mobile devices. In~\cite{Basiri_2016quality}, the authors propose to find errors in OSM data by analyzing patterns extracted from GPS positions and OSM mapping information. For instance, indoor corridors wrongly labeled as tunnels in OSM can be detected by verifying if the trajectory data comes from a pedestrian or a car.

In~\cite{Basiri_2016dataentry}, the authors propose to use GPS positions and machine learning models for recomending the addition of new objects to OSM. For instance, GPS positions can help us predict a missing street in OSM by observing a linear shaped agglomeration of points at some location, where there is not a street in OSM but it is close to the road network. Analyzing the spatiotemporal GPS positions one can also identify, for example, that a road in OSM is a motorway because of the high velocity of objects derived from the GPS information data. The authors in~\cite{Basiri_2016dataentry} extract several features from GPS information, such as the density of nodes in the trajectory and speed of movement. Two types of classifiers are trained: one to predict the geometry and the other to predict object attributes, such as motorway, bicycle lanes, one or two-way street. For geometry classification, the {K-Nearest Neighbours (KNN)} classifier performed better than other algorithms like Logistic Regression and Random Forest. It is observed that the KNN model obtains poor results when detecting polygonal geometries, because of the lack of data along the boundaries of polygonal objects. For the classification of geographical object attributes, the Random Forest classifier outperforms the compared traditional machine learning methods.

In~\cite{Kuntzsch_2016}, the authors utilize GPS information for the reconstruction of road network geometries. The extracted road networks from updated GPS locations can be useful to improve the OSM map. One issue is obtain accurate road geometries since the GPS locations present errors in the range of 5-20 meters. This shortcoming can be mitigated by using multiple trajectories obtained from the same road segment. The authors observe that the accuracy of geometries increase with the number of GPS samples for each road segment.

\subsection{Semantic tags}
\label{subsec:semantic_tags}

The annotation of a geographical object in OSM consists in the digitization of the object geometry (e.g., polygons, lines, or points) and also the attribution of a tags to it. OSM does not provide a rigorous classification system of the geographical objects. It just gives some recommendations and a set of predefined tags that can be used to annotate objects. Thus, the final label attributed to the OSM objects is defined by the volunteers based on their knowledge about the objects under annotation. This can lead to incorrect tag annotations since sometimes it is difficult for inexperienced users to differentiate between similar classes. The decision if a water body is a lake or a pond will depend on the knowledge of the volunteer and on his/her analysis of the aerial imagery or in-situ information. 

\vspace{5mm}
\noindent{\textbf{Recommendation systems}}\\

OSM also allows the assignment of tag values that are not in the set of recommended OSM tag values, which detriments standardization of OSM data. The authors in~\cite{Majic_2017} propose a method to identify recommended OSM tags that are equivalent to new tags created by annotators. The paper proposes an unsupervised method that uses tag usage statistics and geometry type information to compute a similarity measure between a given tag value and a set of common tag values recommended by OSM. This approach just uses OSM data, in contrast with the method proposed in~\cite{Ballatore_2013}, which uses external data, such as information from the OSM wiki website. Other works~\cite{Karagiannakis_2015osmrec,Vandecasteele_2015improving} have implemented tag recommendation tools as plugins of JOSM~\footnote{https://josm.openstreetmap.de/}, a widely used editor of OSM data. In~\cite{Karagiannakis_2015osmrec}, the authors proposed the tool called OSMRec that uses geometrical and textual features to train a Support Vector Machine classifier that is used to recommend a set of tags for new objects that are being digitized by annotators. The tool OSMantic is proposed in~\cite{Vandecasteele_2015improving} and uses semantic similarity and tag frequency to recommend tags.

\vspace{5mm}
\noindent{\textbf{Tags verification}}\\

\begin{figure}[!t]
\begin{center}
\begin{tabular}{c} 
  \includegraphics[width=1.0\columnwidth]{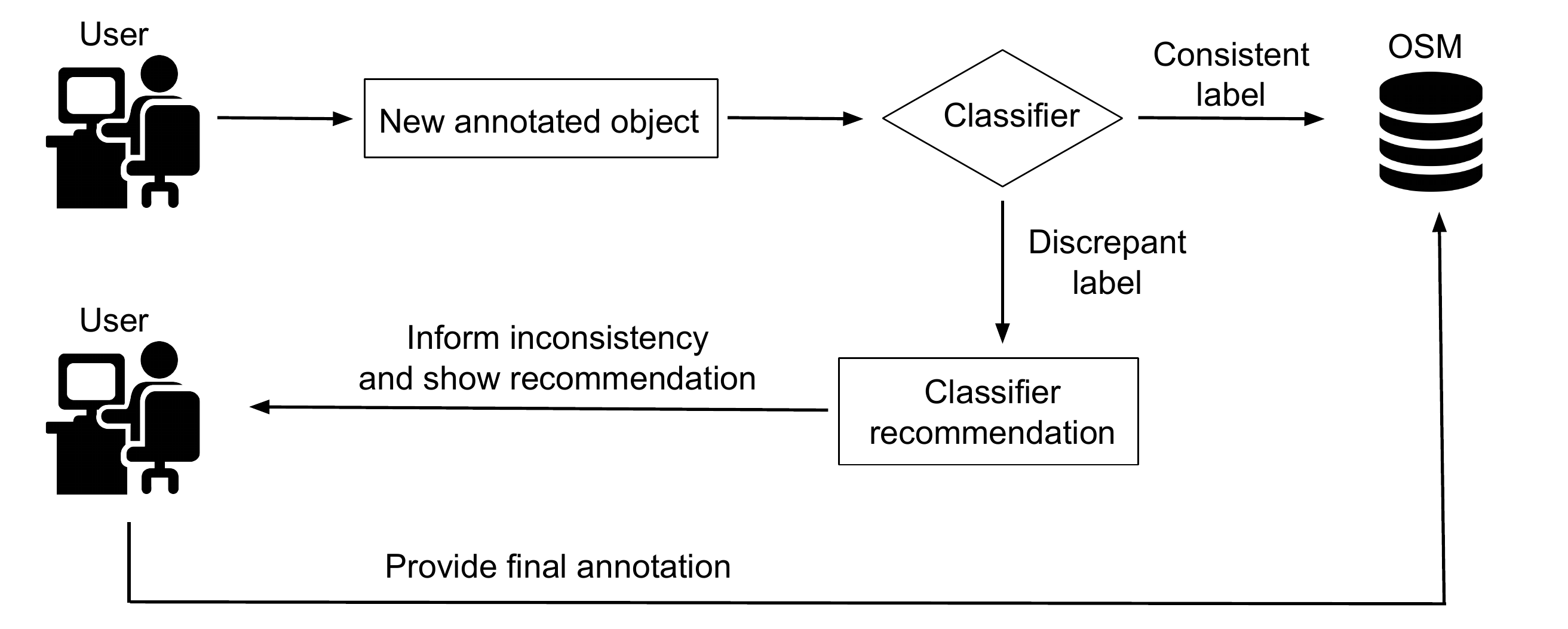} \\
  a) \\
  \includegraphics[width=1.0\columnwidth]{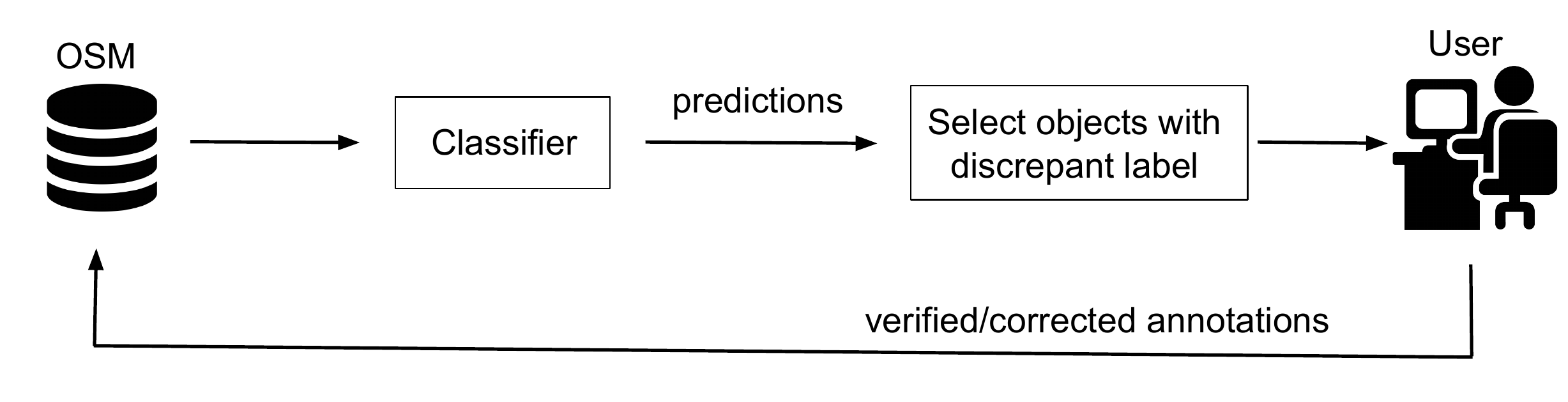} \\
  b) \\
  \includegraphics[width=1.0\columnwidth]{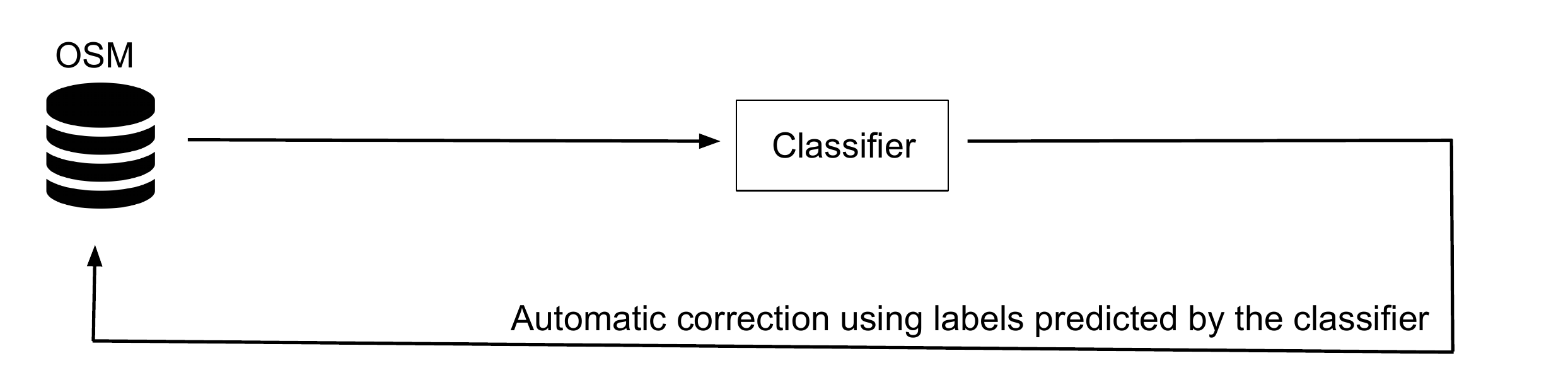} \\
  c)
\end{tabular} 
\end{center}
\caption{Three strategies used with a classifier to improve OSM data, according to~\cite{Ali_2014}. a) \textit{contribution checking}, b) \textit{manual checking} and c) \textit{automatic checking}
	\label{fig:classifier_strategies}}
\end{figure}

As shown in the previous section, several methods propose machine learning models based on properties of OSM objects. Then, the trained classifiers can be applied to another set of OSM objects to find possible annotation errors. In~\cite{Ali_2014}, the authors present three strategies of how such learned classifiers can be applied (see an illustration of them in {Fig.}~\ref{fig:classifier_strategies}). 

\begin{enumerate}
\item \textit{Consistency checking}, where the classifier is applied while the user is editing and assigning tags to OSM objects. In this case, the editing tool can, for example, inform the volunteer that the assigned tag value is inconsistent with what the classifier predicted. Then, the annotator can modify the annotation if required, by taking into consideration the classifier's recommendation
\item \textit{Manual checking}, in which the classifier is applied over a selected set of objects already registered in OSM. Then, the objects whose tags present inconsistencies with the predictions of the classifier are manually validated by the users
\item \textit{Automatic checking}, in which a classifier is used to automatically correct annotations based on its predictions without human verification.
\end{enumerate}

\begin{figure}[!t]
\begin{center}
\begin{tabular}{cc} 
  \includegraphics[width=0.4\columnwidth]{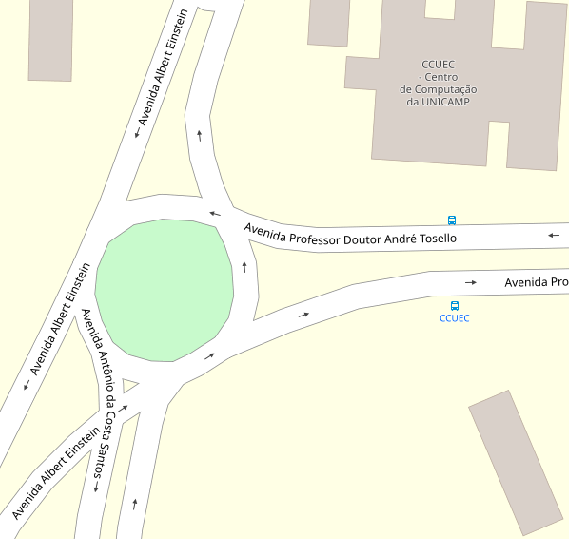} & 
  \includegraphics[width=0.4\columnwidth]{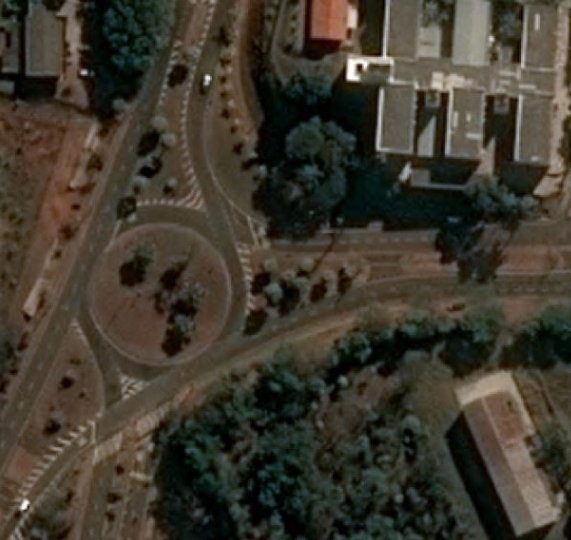} \\
  {a)}  & {b)}  \\  
  \includegraphics[width=0.4\columnwidth]{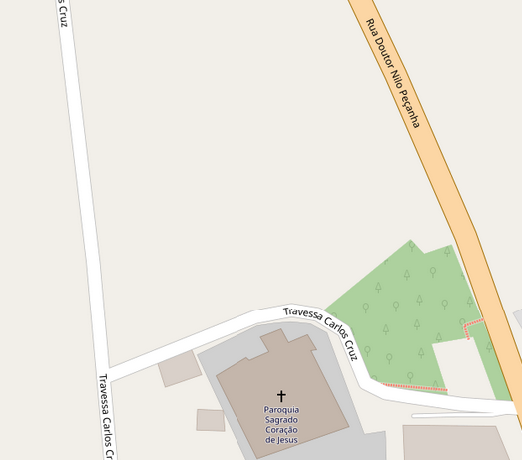} &
  \includegraphics[width=0.4\columnwidth]{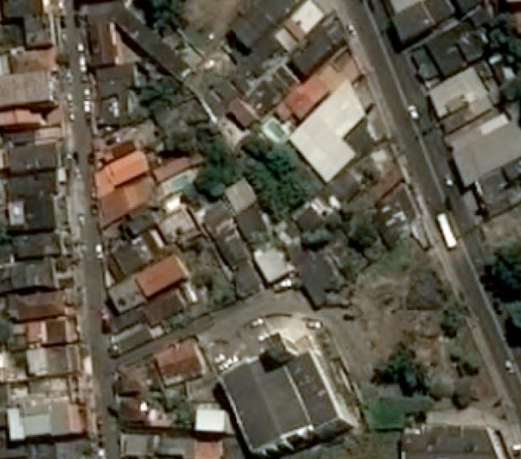} \\ 
  {c)}  & {d)} \\
\end{tabular} 
\end{center}
\caption{Misannotated green area objects in the OSM map alongside with the Bing imagery of the corresponding location: a-b) A grassland area at the center of a roundabout mislabeled as a park in the city of Campinas, Brazil, c-d) A grassland area with some trees (located at the bottom right of the image) is mislabeled as a forest in the city of S\~{a}o Gon\c{c}alo, Brazil.
	\label{fig:green_areas}}
\end{figure}

The method proposed in~\cite{Ali_2014} aims to find errors in tags used for annotating green area objects (i.e., meadow, garden, grass and park). The authors observed that these four types of green area objects are some times mislabeled by OSM annotators. Figures~\ref{fig:green_areas}a-b illustrate a case where a grassland area in the center of a roundabout was wrongly labeled as a park. Figures~\ref{fig:green_areas}c-d depict a case where a grassland area with some trees is wrongly labeled as a forest. The technique proposed in~\cite{Ali_2014} extracts geometrical, topological, and contextual properties (e.g., object area and features based on the 9-Intersection model~\cite{Egenhofer_1995}) and trains a {KNN} classifier to analyze the labels of the four types of green area objects in OSM. The authors in~\cite{Ali_2014} perform an experiment that consists in asking users to manually verify/correct objects with possible erroneous labels. These objects are detected by a classifier, and the experiment shows the effectiveness of the approach to detect mislabeled green areas. 
Another approach proposed in~\cite{Ali_2017rule} tries solving this problem (disambiguation of green areas with the same four classes) by extracting rules from the OSM dataset using the algorithm proposed in~\cite{Agrawal_1994}. These rules are extracted based on topological relations between geographical objects.

\subsection{Points of Interest}
\label{subsec:new_pois}

Points of Interest (POIs) are key elements in OSM. They indicate the location of geographical objects that are commonly used in the city, such as bus stations, cafes, restaurants, ATMs, etc. Thus, the quality control of new added POIs to the OSM database is very important. Some OSM editors, like JOSM implement basic rules to avoid errors (e.g., duplicate elements) while editing objects in OSM. However, this type of topology quality control verifications does not take into account the spatial relationship between a new POI and neighboring geographical objects in OSM.

\vspace{5mm}
\noindent{\textbf{Plausibility of new additions}}\\

In~\cite{Kashian_2019}, the authors propose a recommendation tool that evaluates the positional plausibility of a new registered POI with a certain category label. That work proposes to use spatial co-existence patterns for computing a plausibility score. Confidence scores are computed based on the frequency of occurrence of each pair of POI categories (i.e., \{ATM, Bar\} or \{Bank, supermarket\}). 
Then, the similarity score of two POIs is defined based on the confidence values of their POI categories. Finally, the plausibility score of a new POI is computed as the sum of the similarity score of the new POI and its neighbors. In order to compute the confidence values of pairs of POI categories, the authors recommend using POIs of the same city where the tool has to be validated. This is because different cities may have different patterns.

In~\cite{Kashian_2019}, a case study is shown by evaluating the plausibility of a new ATM being added to four locations in Paris (besides a river and bridge, Paris downtown, middle of a park, and outside the city). The plausibility values obtained are coherent with what is expected -- e.g., the plausibility score of a new ATM located in Paris downtown is much larger than the other alternatives.

\vspace{5mm}
\noindent{\textbf{Tags prediction}}\\

In~\cite{Funke_2017} the authors propose a method that can predict tags of Points of Interest (POIs) based on their names. This method can be useful to extract tag information for POIs that lack of tagging information. For instance, a POI with name ``Chicken Palace" probably should have the tag ``amenity=restaurant". This work used the number of occurences of k-grams, substrings of a given size, extracted from the POI names to create feature vectors. Then, a Random Forest classifier was trained with OSM reference data to food, shop and tourism related POIs in OSM obtaining accurate prediction for some food related classes.

\section{Using OSM data with machine learning algorithms}
\label{sec:osm_applications}

This section presents methods based on machine learning to use OSM in other applications, namely {land use}/
{land cover} classification, building detection and segmentation, navigation, traffic estimation, and fine-scale population estimation.

\subsection{Land use and land cover}
\label{subsec:landuse_and_landcover}

{Land cover}/{land use} mapping has been attempted by governmental organizations (e.g., Urban Atlas~\footnote{https://land.copernicus.eu/local/urban-atlas}), commercial services (e.g., Google maps) and crowdsourced projects (e.g., OpenStreetMap). Several governmental surveys are freely available. However, the quality of the {land use} maps depends on the city and country and also this data is of few use when is not updated frequently. Commercial services like Google maps are more frequently updated but great part of the geographical information in such services are not openly available. In contrast, crowdsourced projects, like OSM provide access to all the collected geographical information and they are regularly updated in several cities.

The data quality of some {land use} types in OSM is comparable to governmental surveys. In~\cite{Assessment_2015arsanjani}, the authors compare the accuracy of OSM data for {land use} mapping in seven large European metropolitan regions. The thematic accuracy and degree of completeness of OSM data are compared to the available Global Monitoring for Environment and Security Urban Atlas (GMESUA) datasets. Evaluation of several land use types suggests that some OSM classes have good quality, such as forest, water bodies, and agricultural areas, and could be used for {land use} planning.

\begin{figure*}
    \centering
    \includegraphics[width=.8\linewidth]{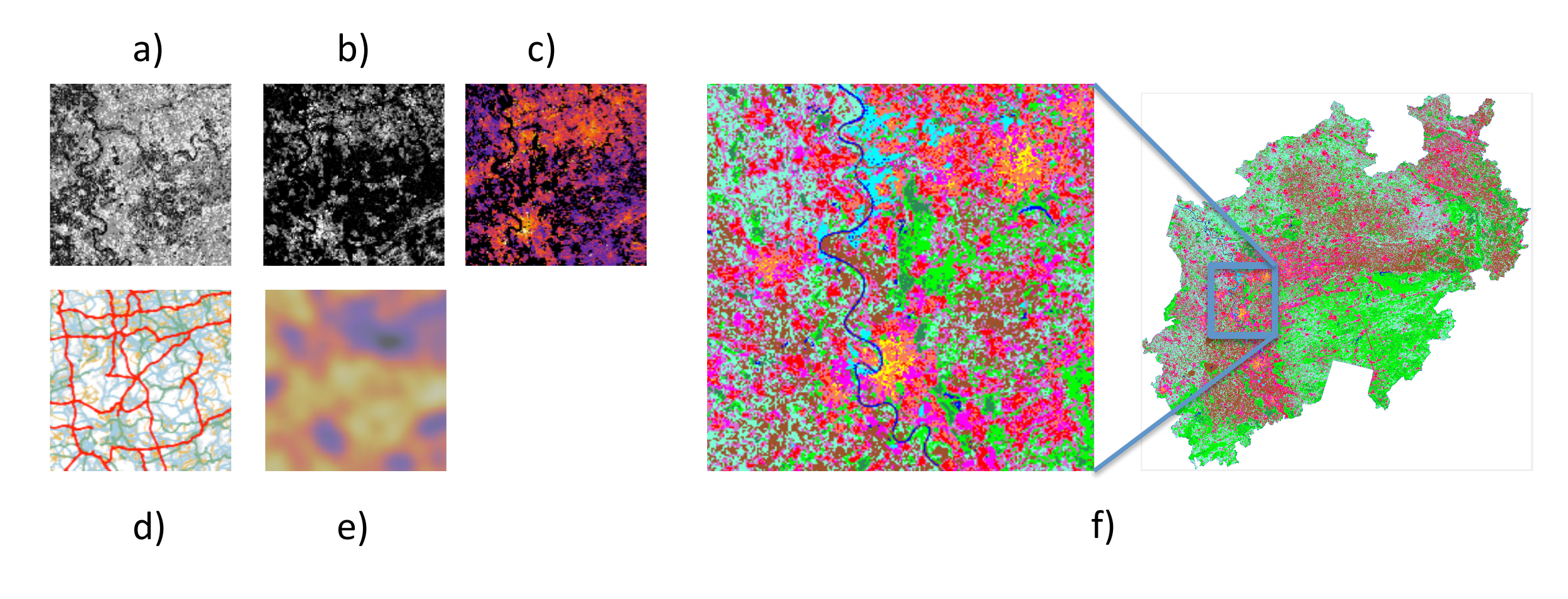}
    \caption{Data sources and local climate zones prediction map in the region of North Rhine Westfalia~\cite{Tuia_2017landusenorth}. All the subfigures correspond to the region in the blue square in the right most map.}
    \label{figWF}
\end{figure*}

Several works have proposed methods to predict 
{land cover}/{land use} labels by using remote sensing imagery and OSM {land use} labels as reference data to train a classifier. The authors in~\cite{Integrating_2016johnson} use time-series Landsat imagery and OSM annotations (i.e., object boundary delineations and {land use} labels) to train and evaluate several supervised methods for 
{land cover} classification, considering six classes (e.g., impervious, farm, forest, grass, orchard, and water). The authors in~\cite{Kaiser_2017} use aerial imagery and a large amount of building and road annotations from OSM as training data for supervised classification. The collected annotations are selected without any quality verification and thus the authors observe several cases of low-quality annotations. The authors show that Convolutional Neural Network (CNN) methods trained with this type of data can achieve high accuracies as compared with methods that use a relatively small amount of good quality data. The authors in~\cite{Audebert_2017} propose a CNN based method that combines aerial imagery and rasterized OSM data to predict per-pixel {land use} labels. In~\cite{Tuia_2017landusenorth}, The authors propose a method for {land use} mapping at large scale (approx. 34000 $km^2$ in the region of NorthRhine-Westfalia, Germany). Each spatial unit is a 200m x 200m cell characterized by multimodal features: remote sensing data (RapidEye bands and texture features, see Fig.~\ref{figWF}a-b), 3D models (see Fig.~\ref{figWF}c) and OSM data (locations of Point of interests and road networks (see Fig.~\ref{figWF}d), both encoded as street densities at the cell level (see Fig.~\ref{figWF}e)).

\begin{figure}[t]
\centering
\includegraphics[width=1.0\columnwidth]{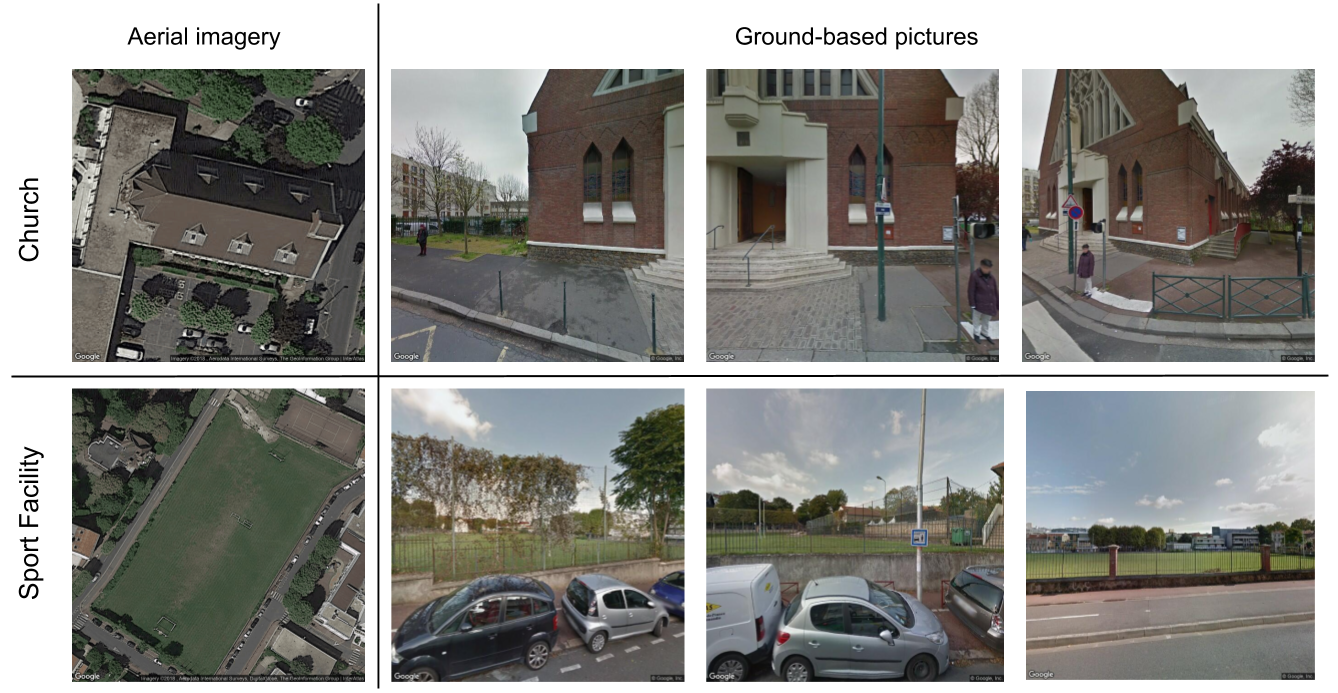}
\caption{Aerial imagery, from Google maps, and ground-based pictures, from Google Street View, for two different OSM objects in the city of Paris. The pictures in the first row correspond to a church. The pictures in the second row correspond to sport facility.\label{fig:aerial_and_gsv_pics}}
\end{figure}

Other works use OSM data and ground-based pictures of a set of OSM objects, to teach a model how to predict the {land use} (e.g., museums, parks, educational institutions, sports centers, and hotels) of other OSM objects~\cite{Srivastava_2018agile,Srivastava_2018,Srivastava_2019understanding}. The method proposed in~\cite{Srivastava_2018agile} use pictures, obtained from Google Street View (GSV), which capture multiple viewpoints of OSM objects (see the last three columns in {Fig.}~\ref{fig:aerial_and_gsv_pics}) and use a pre-trained CNN model to extract features and perform label prediction of 13 {land use} types. The method proposed in~\cite{Srivastava_2018} presents an extension of~\cite{Srivastava_2018agile} and considerably improves the prediction accuracy, {by training a CNN with multiple ground-based pictures, extracting and combining features to identify the proper {land use} class (among 16 {land use} categories) of an OSM object.} 

More recently, the authors in~\cite{Srivastava_2019understanding} propose a CNN-based method that combines aerial imagery and ground-based pictures information to perform {land use} prediction, using OSM as reference data. That method greatly improves the accuracy obtained by the method in~\cite{Srivastava_2018} which uses only ground-based pictures. Additionally, that work also proposes a strategy to deal with the cases when ground-based pictures are missing for an OSM object. Figure~\ref{fig:aerial_and_gsv_pics} illustrates aerial and ground-based images corresponding to two OSM objects in the city of Paris. The first row shows the images of a church. If we just observe the aerial imagery, it is difficult to be confident in predicting the object as a church. Ground-based pictures can give additional visual features to predict the correct {land use} label. The second row shows images of a sport facility in OSM. Because of the fences around the building, sometimes it is difficult to recognize a sport facility. In this case, aerial imagery is usually more valuable to predict the correct {land use} label.

\subsection{Building detection and segmentation}
\label{subsec:building detection}

Building annotations are widely available in OSM and they are usually of considerable quality in several urban areas. Thus, OSM building annotations have been used as reference data to train CNN-based building segmentation methods~\cite{Maggiori_2017cnn, Mnih_2012, Yuan_2018}. {In~\cite{Maggiori_2017recurrent} the authors use a Recurrent Neural Network (RNN) based method that improves the building classification map in several iterations}. The building segmentation maps produced by these methods are not directly usable for Geographical Information Systems (GIS) because they are raster images. However, some methods~\cite{Marcos_2018cvpr,Tasar_2018} have been recently proposed to output vectorial building polygons suitable for GIS software. The method proposes in~\cite{Marcos_2018cvpr} uses Active Contour Models with its parameters learned by a CNN to output vectorial footprints. The authors of~\cite{Maggiori_2017polygonization} and~\cite{Tasar_2018} propose to convert binary building classification maps into vectorial outputs by using a mesh-based approximation method.

In contrast to urban buildings, rural buildings are sparsely located in large geographical areas and their annotations in OSM are less frequent and of lower quality as compared to buildings in urban areas. Some techniques have proposed to improve the detection of locations of rural buildings in aerial imagery~\cite{Chen_2017, Chen_2018}. In~\cite{Chen_2017}, the authors propose a CNN-based method to detect buildings in aerial imagery using an iterative process, in which new samples are selected for annotation by an active learning method and the model is retrained. In~\cite{Chen_2018} the authors use multiple sources of crowdsourced geographical data (namely OSM, MapSwipe~\footnote{https://mapswipe.org/}, and OsmAnd~\footnote{https://osmand.net/}) and an active learning strategy to train a CNN model that detects image patches with buildings. Furthermore, the authors perform an experiment in MapSwipe (smartphone-based application for humanitarian mapping) asking the volunteers to just verify the tiles that are selected by a trained classifier, saving considerably the user effort and obtaining an accurate classifier.

\subsection{Navigation}
\label{subsec:navigation}

In several urban areas, building and road network data in OSM have similar quality as commercial map services. Thus, OSM has been used for navigation/routing applications~\cite{Luxen_2011, Graser_2016, Hahmann_2018}. In~\cite{Luxen_2011}, the authors show that routing services based on OSM data can attain real-time shortest path computation in large areas for web-based applications as well as in hand-held devices. In~\cite{Graser_2016}, the authors show that OSM data can be used for pedestrian routing. The paper proposes a solution to efficient routing in open spaces (e.g., squares, parks, and plazas). A more recent and extensive analysis of different strategies to deal with routing that consider open spaces is presented in~\cite{Hahmann_2018}.  

In~\cite{Floros_2013}, the authors propose a method for accurate global vehicle localization. That work uses visual odometry and OSM data obtaining better localization results than methods that just rely on visual odometry approaches. A probabilistic model for autonomous robot navigation is proposed in~\cite{Suger_2017}. It uses a Markov-Chain Monte-Carlo method to combine semantic terrain information extracted from 3D-LiDAR data and OSM information. 

\subsection{Traffic estimation}
\label{subsec:traffic_estimation}

Traffic prediction is a challenging task that can be very useful for congestion management and car navigation. The authors in~\cite{Xu_2016cross} propose a method to predict four classes of traffic (i.e., good, slow, congested, and extremely congested) in four cities in China. That work uses data obtained from POIs of Baidu maps and geographical objects from OSM (the number of POIs in OSM is limited in China). This geographical data together with other features, like weather, temperature, and house pricing are used to train a machine learning model for traffic prediction. The authors use traffic data from Baidu maps as reference data to train a Support Vector Machine (SVM) classifier. In that work, the authors observe that traffic congestion data is very unbalanced, because most of the time the traffic is good considering all the time intervals in one day (with exception of the rush hours 9h and 17h). Thus, the authors assign higher weights to the classes with less number of samples. The results show that even using class weighting, the accuracy of the model is high for the class good and poor for the other classes. Additionally, that method is compared to the traffic prediction system of Baidu maps outperforming it in some time intervals of the day and obtaining reasonably good performances when the model trained in one city is used for prediction in another city.

In~\cite{Lin_2018transfer}, the authors deal with a similar problem, predicting traffic speed using a regression method. A public dataset named UIUC New York City Traffic Estimates~\footnote{https://lab-work.github.io/data/} is used for their experiments. This dataset contains hourly average traffic speed measurements in the New York City road network, obtained from car trips. Several geographical features are obtained from OSM, such as road length, number of roads connected to the analyzed road, number of neighbor nodes and roads in the area, and also temporal features (e.g., time and whether it is a workday). In order to predict traffic speed in the target areas, the authors of that work propose the method Cluster-based Transfer Model for Prediction (CTMP), which first clusters the road features of the source and target areas. Then, the traffic speed of the target area is computed based on the nearest neighbor roads data of the source area, which contains traffic speed information. CTMP shows better results than other baseline methods, such as Neural Networks and Support Vector Regression. More recently in~\cite{Ren_2019_deep_road}, the authors propose a deep learning method that models the road network topology to predict traffic flow in the city of Chengdu, China. The authors use neural networks to model road network topology and residual learning~\cite{He_2016deep} to model spatio-temporal dependencies. One limitation of that work is that it requires traffic flow historic reference data of the target location to be able to predict the traffic flow in a different time interval.

\subsection{Fine-scale population estimation}
\label{subsec:population_estimation}

Population distribution at the building level is important for several tasks, like urban planning and business development. Population estimation at the building level scale can be obtained by areal interpolation. Although this technique usually requires 3D building models, obtained by LiDAR data, the authors in~\cite{Bakillah_2014fine} propose to use building footprints and POIs from OSM to predict population distribution by using areal interpolation. The authors in~\cite{Yao_2017mapping} use a Random Forest model to predict population at the grid level (i.e., the area of study is divided into grid cells) by using Baidu POIs, mobile user density data, and road networks from OSM. Then, the grid level estimations are transformed to building level estimations.

In~\cite{Gervasoni_2018convolutional}, the authors propose a CNN-based method to perform population density estimation at the grid level. First, the area of study is divided into grid cells of size 200 $\times$ 200 meters. Then, for each cell, they compute several urban features, such as building area, number of buildings, and number of POIs. Finally, the authors use a fully convolutional neural network, applied over the urban features of the grid cells, to obtain the population estimation of the corresponding grid cells. The experimental results show that by training the model with data from 14 French cities, the model attains low error rates in the validation data extracted from the city of Lyon.

\section{Discussion and future works}
\label{sec:future_works}

In this section, we discuss potential promising research avenues at the interface of OSM and machine learning.

\vspace{5mm}
\noindent{\textbf{{Applications with multimodal data}}}\\

Although the use of several sources of data {(multimodal approaches)} have proven to be beneficial for solving several problems it has been only applied in relatively few works described in this manuscript, such as in~\cite{Srivastava_2019understanding}. We believe that the performance of supervised methods used to improve OSM data can be greatly improved with the use of several data sources, such as images, tracking data, and social media data. This has been also pointed out in~\cite{Touya_2017assessing} for the particular case of POIs, where the authors recommend the use of several data sources, like OSM geographical data, ground-based pictures, and historical data to create more accurate models for POI label prediction. Among these external data sources, ground-based pictures obtained from Google Street View (GSV) have found to be particularly useful to enrich OSM data, for example, for {land use} prediction~\cite{Srivastava_2018} and crosswalk localization~\cite{Berriel_2017automatic}. Recently, crowdsourced ground-based images collected by the TeleNav's project, called OpenStreetCam\footnote{https://openstreetcam.org}, has also been used to improve OSM. For instance, pictures obtained from OpenStreetCam have been used to detect traffic signs\footnote{https://blog.improveosm.org/en/2018/02/detecting-traffic-signs-in-openstreetcam/}. {Similarly, Mapillary has released a dataset of ground-based pictures\footnote{https://www.mapillary.com/dataset/trafficsign} for traffic sign detection~\cite{Ertler_2019traffic}}. Although nowadays the coverage of crowdsourced ground-based pictures platforms is not as complete as GSV, they have the potential to obtain more updated data and to be available for everyone at no cost. This data can then be used to obtain up-to-date OSM data of objects like traffic lights and road signs, which can greatly benefit navigation applications.

{Very recent works~\cite{Haberle_2019geo, Haberle_2019building} have explored the use of OSM data combined with social media textual information, obtained from Twitter, to perform land use classification. These works use natural language processing techniques (e.g, fastText~\cite{Bojanowski_2017enriching}) together with neural networks to process geolocalized tweets. The authors in \cite{Huang_2018classification} use Long short-term memory (LSTM) for feature extraction of Twitter data for the classification of OSM urban-objects into three categories: residential, non-residential and mixed-use. In contrast to works that use image data for finegrained land use classification~\cite{Srivastava_2019understanding}, these works present results with datasets that contain only a few land use classes, but have the advantage of the massive amounts of data that can be retrieved from social media. Also, working with Twitter, it creates the need of specialized workflows, typically aiming at denoising data, retrieve useful Tweets, or avoiding spourious geolocation due to re-Tweets. We found these issues very exciting and believe that textual data should be used in a multimodal fashion: as suggested in~\cite{Haberle_2019geo}, textual information and remote sensing data could be used together to perform more precise land use classification and this is, in our opinion, one of the avenues that future research could focus on in the next years.}

\vspace{5mm}
\noindent{\textbf{{Multimodal techniques}}}\\

{In~\cite{Srivastava_2019understanding}, the authors use a CNN that extracts features from ground-based pictures and remote sensing images to perform land use classification of OSM urban objects. The features extracted from these two different image modalities are stacked and a fully connected layer is used to predict the land use category. The authors in~\cite{Srivastava_2019understanding} also propose a solution to the case when the data of one of the modalities is missing. This is done by using Cannonical Correlation Analysis (CCA) to create a latent space and projection matrices that allow, starting from a satellite patch, to retrieve a plausible feature vector for the ground-based missing images.  
Another example of feature stacking, but using governmental data, is proposed in~\cite{Workman_2017unified}. In that work, the authors propose a method for per-pixel land use classification. The proposed method stacks features extracted from remote sensing data and ground-based pictures, but keep the spatial information. Finally, a Multilayer Perceptron (MLP) is used to perform per-pixel classification.
In~\cite{Zhang_2019fusion}, the authors perform local climate zones classifications by fusing multi-temporal and multispectral satellite images that have different spatial resolutions. The authors propose a method for weighted voting of the output of the classifiers trained with the different data modalities and show more accurate results than feature stacking. The methods that we have commented in this review perform the fusion of data of two modalities. We believe that a challenging topic of future research is the creation of methods that handle the fusion of more than two modalities, dealing well with the problem of missing modality. 
}  

\vspace{5mm}
\noindent{\textbf{Supporting users via  interaction and skills estimation}}\\

Mapping information obtained by machine learning models applied with Earth Observation data (e.g., building footprints, and road networks) could contain some errors. Thus, an alternative to performing automatic updates in OSM is to use a human-computer interactive approach. In this strategy, the machine learning model is used to minimize the effort of the users during the annotation process. This strategy has been already applied in~\cite{Bastani_2018maid} to improve road network completeness. In that work, an automatic method is used to extract major roads (in places with a few road annotations) and missing roads (in places where major roads are already annotated) and the user is asked to verify or correct if needed the extracted roads. This work shows experimentally that such an interactive approach is more efficient than traditional manual annotation. A related and also effective approach that involves the user in the process is active learning~\cite{Settles_2009active,Cra12}. Active learning is an iterative process that consists of intelligently selecting a small number of samples for user annotation that allow training an effective classifier. 
We think that these approaches that involve the interaction between a machine learning model and the annotator could be applied to improve other aspects of OSM data with fewer annotators' effort, especially if user's skills are involved in the process, as shown in concurrent research in crowdsourcing~\cite{GOm11,Tui12c}. {Although active learning methods have been applied to OSM data of rural areas~\cite{Chen_2017, Chen_2018}, they can be also  applied to OSM data of urban areas (e.g., cities in developing countries).}

Between January and May of 2019, the number of active contributors per month in OSM was less than 1\% of the total registered OSM users. However, a few efforts have been made to encourage volunteers to frequently edit data. Gamification strategies could be applied to solve the problem, by assigning annotation tasks to volunteers with a game-like interface and scoring systems. Some gamification projects have been listed in the OSM wiki~\footnote{https://wiki.openstreetmap.org/wiki/Gamification} but they are not of widespread use.

Contributor analysis in OSM have been studied in several works~\cite{Neis_2012analyzing}. It has been observed that volunteers' experience and familiarity with the area edited in the map are good proxies to estimate the quality of their annotations in OSM. The authors in~\cite{Severinsen_2019vgtrust} propose a measure for estimating annotation trust, by using annotation statistics obtained from volunteers' activities, object geometries, and temporal data. A trust index could be also learned with machine learning methods using OSM data statistics with some reference data. Such trust index could be used to improve methods that are created to verify the correctness of labels of objects registered in OSM~\cite{Ali_2014, Ali_2017rule}.

\vspace{5mm}
\noindent{\textbf{New ways of searching and interacting with OSM}}\\

\begin{figure}[t]
\centering
\includegraphics[width=1.0\columnwidth]{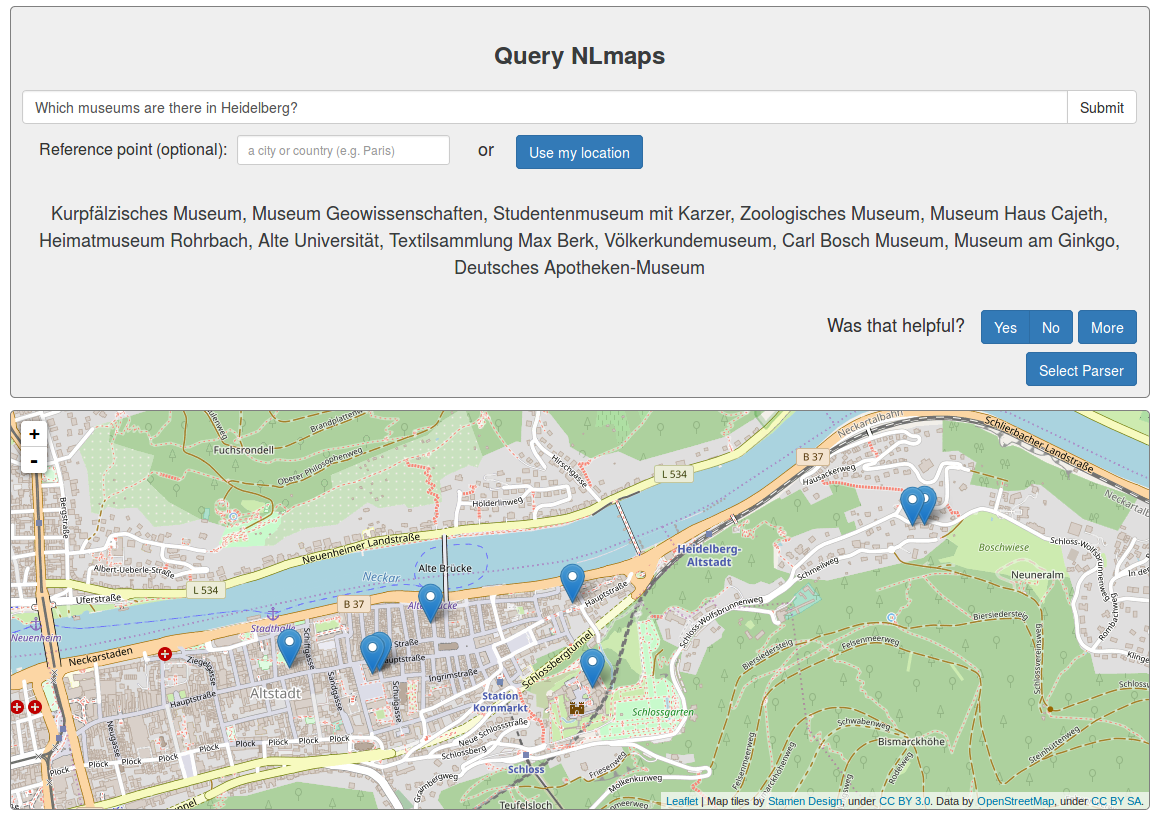}
\caption{NLmaps search results for the query: ``Which museums are there in Heildelberg?''.\label{fig:nlmaps_query}}
\end{figure}

Search tools are important features of mapping services. However, the search tool provided by the OSM website has limited capabilities, basically just trying to find the location of a given place name or address. In~\cite{Lawrence_2016nlmaps}, the authors propose NLmaps~\footnote{https://www.cl.uni-heidelberg.de/statnlpgroup/nlmaps/}, a natural language interface to query data in OSM. This service can answer textual questions about geographical facts in OSM. The response is a text and a map with geographical objects of interest highlighted. For instance, NLmaps can answer the following question: ``What is the closest supermarket from the Royal Bank of Scotland in Edinburgh?". Figure~\ref{fig:nlmaps_query} illustrates the search results in NLmaps for the question: ``Which museums are there in Heidelberg?". First, the names of the museums found are shown and then their locations are shown with markers in the OSM map of the city of Heidelberg.

This work uses a semantic parser for the OSM data, proposed in~\cite{Haas_2016corpus} and extended in~\cite{Lawrence_2018improving}, to transform natural language text to a Machine Readable Language formula (MRL)~\cite{Andreas_2013semantic}. This representation is used to create a structured query and retrieve OSM information using an extension of the Overpass API, called Overpass MLmaps~\footnote{https://github.com/carhaas/overpass-nlmaps}. Although NLmaps can handle several common questions, it is not able to answer complex questions like ``Where are 4 star hotels in Berlin?". The authors of~\cite{Lawrence_2018improving} also propose to improve NLmaps performance by using user feedback. A simple example of feedback is to ask the user if the result of the query is helpful or not. More complex feedback can be obtained from expert users, with knowledge of OSM and the Overpass API. For instance, one can ask the expert user if the intermediate results of the query processing pipeline are correct or not. We believe that the implementation of a more intelligent tool for answering natural language queries in the OSM website could potentially attract more users.

Recently, the authors of~\cite{Lobry_2019} use OSM data for the Visual Question Answering (VQA) problem, a task that consists in providing a natural language answer for a given image and a natural language question about the image. The authors create a dataset using OSM data and Sentinel-2 satellite imagery for training a VQA model. VQA is a challenging problem and its application for remote sensing imagery is still in its infancy. VQA can enable to perform some queries in places where there is incomplete mapping data, but aerial imagery is available.

\vspace{5mm}
\noindent{\textbf{OSM-enabled augmented reality}}\\

Recent applications of OSM data can be further improved by using machine learning methods. For instance, augmented reality (AR) has great potential to improve the way we experience the cities, especially for tourism. In~\cite{Ruta_2014semantic}, an augmented reality tool is presented to recommend Points Of Interest (POIs) in the city by taking into account the profile of the users. The authors propose a mobile application that shows POIs registered in OSM in real time with the mobile camera view as background. A case study in the city of Trani, Italy, shows how a tourist interested in local architectural work can visualize in the mobile application the POIs, within a distance radius, marked as colored circles with labels superimposed over the mobile camera view. In~\cite{Ruta_2014semantic}, a hand engineered rule is proposed to define which POIs to show to the user. For this type of AR applications,  reference data could be used by asking users' feedback (e.g., asking to the user to add POIs visited in a favorites list). Thus, obtaining reference data can enable the use of more effective supervised learning methods to predict adequate POIs for new users.

\section{Conclusion}
\label{sec:conclusions}

In this paper, we reviewed synergetic efforts involving OpenStreetMap and machine learning. In the first part, we review works that use machine learning to improve OSM data. These methods deal with the three main object geometric types in OSM: points (e.g., ATMs), lines (e.g., roads) and polygons (e.g., buildings). The reviewed works use frequently traditional machine learning methods (e.g., Support Vector Machines and Random Forest), but also several of them have used state-of-the-art methods, such as deep learning based techniques, especially when dealing with remote sensing image data. Although several methods could be integrated into OSM editors (e.g., iD editor~\footnote{http://ideditor.com/} and JOSM) just a few works~\cite{Karagiannakis_2015osmrec,Vandecasteele_2015improving} have implemented their methods in such tools.

The second part of the manuscript reviewed works that have used machine learning based techniques to use OSM data for applications in other domains. We identified two groups of works. The first group uses OSM data as reference data to train machine learning models, for examples several works that perform {land use} classification~\cite{Srivastava_2018, Srivastava_2019understanding} and building segmentation~\cite{Maggiori_2017cnn, Mnih_2012}. The second group uses OSM data to extract features for training the machine learning model (e.g., fine-scale population estimation~\cite{Gervasoni_2018convolutional}). 

We believe that a mixture of automatic and human-interactive approaches could lead to obtaining accurate data for OSM with efficient use of the annotators' labor. The strong links with machine learning and the ever increasing availability of up-to-date remote sensing data open countless opportunities for research in this exciting interface among disciplines. 
 
\section*{Acknowledgment}
This research was funded by Funda\c{c}\~{a}o de Amparo \`{a} Pesquisa do
Estado de S\~{a}o Paulo (FAPESP, grant 2016/14760-5 and 2014/12236-1), Conselho Nacional de Desenvolvimento Cient\'{\i}fico e Tecnol\'{o}gico (CNPq, grant 303808/2018-7) and Coordena\c{c}\~{a}o de Aperfei\c{c}oamento de Pessoal de N\'{\i}vel Superior - Brasil (CAPES, finance code 001).

\bibliographystyle{IEEEtran}
\bibliography{refs}

\begin{thebibliography}{100}
\providecommand{\url}[1]{#1}
\csname url@samestyle\endcsname
\providecommand{\newblock}{\relax}
\providecommand{\bibinfo}[2]{#2}
\providecommand{\BIBentrySTDinterwordspacing}{\spaceskip=0pt\relax}
\providecommand{\BIBentryALTinterwordstretchfactor}{4}
\providecommand{\BIBentryALTinterwordspacing}{\spaceskip=\fontdimen2\font plus
\BIBentryALTinterwordstretchfactor\fontdimen3\font minus
  \fontdimen4\font\relax}
\providecommand{\BIBforeignlanguage}[2]{{%
\expandafter\ifx\csname l@#1\endcsname\relax
\typeout{** WARNING: IEEEtran.bst: No hyphenation pattern has been}%
\typeout{** loaded for the language `#1'. Using the pattern for}%
\typeout{** the default language instead.}%
\else
\language=\csname l@#1\endcsname
\fi
#2}}
\providecommand{\BIBdecl}{\relax}
\BIBdecl

\bibitem{phd_john}
J.~E. Vargas-Mu{\~{n}}oz, ``Machine learning applied to open geographical
  data,'' Ph.D. dissertation, University of Campinas, 10 2019.

\bibitem{Goodchild_2007}
M.~F. Goodchild, ``Citizens as sensors: the world of volunteered geography,''
  \emph{GeoJournal}, vol.~69, no.~4, pp. 211--221, 2007.

\bibitem{Mooney_2012}
P.~Mooney and P.~Corcoran, ``{The annotation process in OpenStreetMap},''
  \emph{Transactions in GIS}, vol.~16, no.~4, pp. 561--579, 2012.

\bibitem{Haklay_2010}
M.~Haklay, ``{How good is volunteered geographical information? A comparative
  study of OpenStreetMap and Ordnance Survey datasets},'' \emph{Environment and
  planning B: Planning and design}, vol.~37, no.~4, pp. 682--703, 2010.

\bibitem{Funke_2015}
S.~Funke, R.~Schirrmeister, and S.~Storandt, ``{Automatic extrapolation of
  missing road network data in OpenStreetMap},'' in \emph{Proceedings of the
  2nd International Conference on Mining Urban Data-Volume 1392}, 2015, pp.
  27--35.

\bibitem{Jilani_2014}
M.~Jilani, P.~Corcoran, and M.~Bertolotto, ``{Automated highway tag assessment
  of OpenStreetMap road networks},'' in \emph{Proceedings of the 22nd ACM
  SIGSPATIAL International Conference on Advances in Geographic Information
  Systems}, 2014, pp. 449--452.

\bibitem{Xu_2017}
Y.~Xu, Z.~Chen, Z.~Xie, and L.~Wu, ``Quality assessment of building footprint
  data using a deep autoencoder network,'' \emph{International Journal of
  Geographical Information Science}, vol.~31, no.~10, pp. 1929--1951, 2017.

\bibitem{Koukoletsos_2012}
T.~Koukoletsos, M.~Haklay, and C.~Ellul, ``{Assessing Data Completeness of VGI
  through an Automated Matching Procedure for Linear Data},''
  \emph{Transactions in GIS}, vol.~16, 08 2012.

\bibitem{Fan_2014}
H.~Fan, A.~Zipf, Q.~Fu, and P.~Neis, ``{Quality assessment for building
  footprints data on OpenStreetMap},'' \emph{International Journal of
  Geographical Information Science}, vol.~28, pp. 700--719, 04 2014.

\bibitem{Girres_2010}
J.~F. Girres and G.~Touya, ``{Quality assessment of the French OpenStreetMap
  dataset},'' \emph{Transactions in GIS}, vol.~14, no.~4, pp. 435--459, 2010.

\bibitem{Neis_2011}
P.~Neis, D.~Zielstra, and A.~Zipf, ``{The street network evolution of
  crowdsourced maps: OpenStreetMap in Germany 2007--2011},'' \emph{Future
  Internet}, vol.~4, no.~1, pp. 1--21, 2011.

\bibitem{Neis_2014}
P.~Neis and D.~Zielstra, ``{Recent developments and future trends in
  volunteered geographic information research: The case of OpenStreetMap},''
  \emph{Future Internet}, vol.~6, no.~1, pp. 76--106, 2014.

\bibitem{Arsanjani_2015exploration}
J.~J. Arsanjani, P.~Mooney, M.~Helbich, and A.~Zipf, ``{An exploration of
  future patterns of the contributions to OpenStreetMap and development of a
  Contribution Index},'' \emph{Transactions in GIS}, vol.~19, no.~6, pp.
  896--914, 2015.

\bibitem{Senaratne_2017}
H.~Senaratne, A.~Mobasheri, A.~L. Ali, C.~Capineri, and M.~Haklay, ``A review
  of volunteered geographic information quality assessment methods,''
  \emph{International Journal of Geographical Information Science}, vol.~31,
  no.~1, pp. 139--167, 2017.

\bibitem{Jilani_2019traditional}
M.~Jilani, M.~Bertolotto, P.~Corcoran, and A.~Alghanim, ``{Traditional vs.
  Machine-Learning Techniques for OSM Quality Assessment},'' in
  \emph{Geospatial Intelligence: Concepts, Methodologies, Tools, and
  Applications}.\hskip 1em plus 0.5em minus 0.4em\relax IGI Global, 2019, pp.
  469--487.

\bibitem{Fonte_2015}
C.~C. Fonte, L.~Bastin, L.~See, G.~Foody, and F.~Lupia, ``{Usability of VGI for
  validation of land cover maps},'' \emph{International Journal of Geographical
  Information Science}, vol.~29, no.~7, pp. 1269--1291, 2015.

\bibitem{Srivastava_2018}
S.~Srivastava, J.~E. Vargas-Mu{\~{n}}oz, S.~Lobry, and D.~Tuia, ``Fine-grained
  landuse characterization using ground-based pictures: a deep learning
  solution based on globally available data,'' \emph{International Journal of
  Geographical Information Science}, vol.~0, no.~0, pp. 1--20, 2018.

\bibitem{Audebert_2017}
N.~Audebert, B.~L. Saux, and S.~Lef{\`{e}}vre, ``{Joint Learning from Earth
  Observation and OpenStreetMap Data to Get Faster Better Semantic Maps},'' in
  \emph{IEEE Conference on Computer Vision and Pattern Recognition Workshops,
  {CVPR} Workshops 2017, Honolulu, HI, USA, July 21-26, 2017}, 2017, pp.
  1552--1560.

\bibitem{Srivastava_2019understanding}
S.~Srivastava, J.~E. Vargas-Mu{\~n}oz, and D.~Tuia, ``Understanding urban
  landuse from the above and ground perspectives: A deep learning, multimodal
  solution,'' \emph{Remote Sensing of Environment}, vol. 228, pp. 129--143,
  2019.

\bibitem{Lin_2018transfer}
B.~Y. Lin, F.~F. Xu, E.~Q. Liao, and K.~Q. Zhu, ``{Transfer Learning for
  Traffic Speed Prediction: A Preliminary Study},'' in \emph{Workshops at the
  Thirty-Second AAAI Conference on Artificial Intelligence}, 2018.

\bibitem{Schmitz_2008}
S.~Schmitz, A.~Zipf, and P.~Neis, ``New applications based on collaborative
  geodata—the case of routing,'' in \emph{Proceedings of XXVIII INCA
  international congress on collaborative mapping and space technology}, 2008.

\bibitem{Neis_2015}
P.~Neis, ``{Measuring the Reliability of Wheelchair User Route Planning based
  on Volunteered Geographic Information},'' \emph{Transactions in GIS},
  vol.~19, no.~2, pp. 188--201, 2015.

\bibitem{VargasMunoz_2019}
J.~E. Vargas-Mu{\~{n}}oz, S.~Lobry, A.~X. Falc{\~{a}}o, and D.~Tuia,
  ``{Correcting rural building annotations in OpenStreetMap using convolutional
  neural networks},'' \emph{ISPRS Journal of Photogrammetry and Remote
  Sensing}, vol. 147, pp. 283 -- 293, 2019.

\bibitem{Mnih_2012}
V.~Mnih and G.~E. Hinton, ``Learning to label aerial images from noisy data,''
  in \emph{International Conference on Machine Learning}, 2012, pp. 567--574.

\bibitem{Over_2010}
M.~Over, A.~Schilling, S.~Neubauer, and A.~Zipf, ``{Generating web-based 3D
  City Models from OpenStreetMap: The current situation in Germany},''
  \emph{Computers, Environment and Urban Systems}, vol.~34, no.~6, pp.
  496--507, nov 2010.

\bibitem{Goetz_2012}
M.~Goetz and A.~Zipf, ``{Using Crowdsourced Geodata for Agent-Based Indoor
  Evacuation Simulations},'' \emph{ISPRS International Journal of
  Geo-Information}, vol.~1, no.~2, pp. 186--208, 2012.

\bibitem{Schilling_2009}
A.~Schilling, M.~Over, S.~Neubauer, P.~Neis, G.~Walenciak, and A.~Zipf,
  ``{Interoperable Location Based Services for 3D cities on the Web using user
  generated content from OpenStreetMap},'' \emph{Urban and regional data
  management: UDMS annual}, pp. 75--84, 2009.

\bibitem{Zhu17}
X.~Zhu, D.~{Tuia}, L.~Mou, G.~Xia, L.~Zhang, F.~Xu, and F.~Fraundorfer, ``Deep
  learning in remote sensing: A comprehensive review and list of resources,''
  \emph{{IEEE Geosci. Remote Sens. Mag.}}, vol.~5, no.~4, pp. 8--36, 2017.

\bibitem{Neis_2012towards}
P.~Neis, M.~Goetz, and A.~Zipf, ``{Towards automatic vandalism detection in
  OpenStreetMap},'' \emph{ISPRS International Journal of Geo-Information},
  vol.~1, no.~3, pp. 315--332, 2012.

\bibitem{Truong_2018}
Q.~Truong, G.~Touya, and C.~De~Runz, ``{Towards Vandalism Detection in
  OpenStreetMap Through a Data Driven Approach},'' in \emph{GIScience 2018},
  2018.

\bibitem{Zhuo_2018}
X.~Zhuo, F.~Fraundorfer, F.~Kurz, and P.~Reinartz, ``{Optimization of
  OpenStreetMap Building Footprints Based on Semantic Information of Oblique
  UAV Images},'' \emph{Remote Sensing}, vol.~10, no.~4, p. 624, 2018.

\bibitem{Barrington_2017world}
C.~Barrington-Leigh and A.~Millard-Ball, ``{The world’s user-generated road
  map is more than 80\% complete},'' \emph{PloS one}, vol.~12, no.~8, p.
  e0180698, 2017.

\bibitem{Cheng_2017automatic}
G.~Cheng, Y.~Wang, S.~Xu, H.~Wang, S.~Xiang, and C.~Pan, ``Automatic road
  detection and centerline extraction via cascaded end-to-end convolutional
  neural network,'' \emph{IEEE Transactions on Geoscience and Remote Sensing},
  vol.~55, no.~6, pp. 3322--3337, 2017.

\bibitem{Mattyus_2017deeproadmapper}
G.~M{\'a}ttyus, W.~Luo, and R.~Urtasun, ``Deeproadmapper: Extracting road
  topology from aerial images,'' in \emph{IEEE International Conference on
  Computer Vision}, 2017, pp. 3438--3446.

\bibitem{Bastani_2018roadtracer}
F.~Bastani, S.~He, S.~Abbar, M.~Alizadeh, H.~Balakrishnan, S.~Chawla,
  S.~Madden, and D.~DeWitt, ``Roadtracer: Automatic extraction of road networks
  from aerial images,'' in \emph{IEEE Conference on Computer Vision and Pattern
  Recognition}, 2018, pp. 4720--4728.

\bibitem{Ventura_2018iterative}
C.~Ventura, J.~Pont{-}Tuset, S.~Caelles, K.~K. Maninis, and L.~V. Gool,
  ``Iterative deep learning for road topology extraction,'' in \emph{British
  Machine Vision Conference}, 2018.

\bibitem{Bastani_2018maid}
F.~Bastani, S.~He, S.~Abbar, M.~Alizadeh, H.~Balakrishnan, S.~Chawla, and
  S.~Madden, ``Machine-assisted map editing,'' in \emph{Proceedings of the 26th
  ACM SIGSPATIAL International Conference on Advances in Geographic Information
  Systems}, ser. SIGSPATIAL '18.\hskip 1em plus 0.5em minus 0.4em\relax ACM,
  2018, pp. 23--32.

\bibitem{Mattyus_2016hd}
G.~M{\'a}ttyus, S.~Wang, S.~Fidler, and R.~Urtasun, ``{Hd maps: Fine-grained
  road segmentation by parsing ground and aerial images},'' in \emph{IEEE
  Conference on Computer Vision and Pattern Recognition}, 2016, pp. 3611--3619.

\bibitem{Jilani_2013}
M.~Jilani, P.~Corcoran, and M.~Bertolotto, ``{Automated quality improvement of
  road network in OpenStreetMap},'' in \emph{Agile Workshop (Action and
  Interaction in Volunteered Geographic Information)}, 2013, p.~19.

\bibitem{Jilani_2013multi}
------, ``{Multi-granular street network representation towards quality
  assessment of OpenStreetMap data},'' in \emph{Proceedings of the Sixth ACM
  SIGSPATIAL International Workshop on Computational Transportation
  Science}.\hskip 1em plus 0.5em minus 0.4em\relax ACM, 2013, p.~19.

\bibitem{Jilani_2016probabilistic}
------, ``Probabilistic graphical modelling for semantic labelling of
  crowdsourced map data,'' in \emph{Intelligent Systems Technologies and
  Applications}.\hskip 1em plus 0.5em minus 0.4em\relax Springer, 2016, pp.
  213--224.

\bibitem{Tsochantaridis_2005largemargin}
I.~Tsochantaridis, T.~Joachims, T.~Hofmann, and Y.~Altun, ``Large margin
  methods for structured and interdependent output variables,'' \emph{Journal
  of Machine Learning Research}, vol.~6, no. Sep, pp. 1453--1484, 2005.

\bibitem{Jilani_2018multi}
M.~Jilani, P.~Corcoran, and M.~Bertolotto, ``{A Multi-layer CRF Based
  Methodology for Improving Crowdsourced Street Semantics},'' in
  \emph{Proceedings of the 11th ACM SIGSPATIAL International Workshop on
  Computational Transportation Science}.\hskip 1em plus 0.5em minus 0.4em\relax
  ACM, 2018, pp. 29--38.

\bibitem{Jepsen_2019graph}
T.~S. Jepsen, C.~S. Jensen, and T.~D. Nielsen, ``Graph convolutional networks
  for road networks,'' in \emph{Proceedings of the 27th ACM SIGSPATIAL
  International Conference on Advances in Geographic Information Systems},
  2019, pp. 460--463.

\bibitem{Li_2014polygon}
Q.~Li, H.~Fan, X.~Luan, B.~Yang, and L.~Liu, ``{Polygon-based approach for
  extracting multilane roads from OpenStreetMap urban road networks},''
  \emph{International Journal of Geographical Information Science}, vol.~28,
  no.~11, pp. 2200--2219, 2014.

\bibitem{Xu_2019multilane}
Y.~Xu, Z.~Xie, L.~Wu, and Z.~Chen, ``{Multilane roads extracted from the
  OpenStreetMap urban road network using random forests},'' \emph{Transactions
  in GIS}, vol.~23, no.~2, pp. 224--240, 2019.

\bibitem{Mattyus_2015enhancing}
G.~M{\'a}ttyus, S.~Wang, S.~Fidler, and R.~Urtasun, ``Enhancing road maps by
  parsing aerial images around the world,'' in \emph{IEEE International
  Conference on Computer Vision}, 2015, pp. 1689--1697.

\bibitem{Basiri_2016quality}
A.~Basiri, M.~Jackson, P.~Amirian, A.~Pourabdollah, M.~Sester, A.~Winstanley,
  T.~Moore, and L.~Zhang, ``{Quality assessment of OpenStreetMap data using
  trajectory mining},'' \emph{Geo-spatial Information Science}, vol.~19, no.~1,
  pp. 56--68, 2016.

\bibitem{Basiri_2016dataentry}
A.~Basiri, P.~Amirian, and P.~Mooney, ``Using crowdsourced trajectories for
  automated osm data entry approach,'' \emph{Sensors}, vol.~16, no.~9, p. 1510,
  2016.

\bibitem{Kuntzsch_2016}
C.~Kuntzsch, M.~Sester, and C.~Brenner, ``Generative models for road network
  reconstruction,'' \emph{International Journal of Geographical Information
  Science}, vol.~30, no.~5, pp. 1012--1039, 2016.

\bibitem{Majic_2017}
I.~Majic, S.~Winter, and M.~Tomko, ``{Finding Equivalent Keys in Openstreetmap:
  Semantic Similarity Computation Based on Extensional Definitions},'' in
  \emph{Proceedings of the 1st Workshop on Artificial Intelligence and Deep
  Learning for Geographic Knowledge Discovery}, ser. GeoAI '17.\hskip 1em plus
  0.5em minus 0.4em\relax New York, NY, USA: ACM, 2017, pp. 24--32.

\bibitem{Ballatore_2013}
A.~Ballatore, M.~Bertolotto, and D.~C. Wilson, ``{Geographic knowledge
  extraction and semantic similarity in OpenStreetMap},'' \emph{Knowledge and
  Information Systems}, vol.~37, no.~1, pp. 61--81, 2013.

\bibitem{Karagiannakis_2015osmrec}
N.~Karagiannakis, G.~Giannopoulos, D.~Skoutas, and S.~Athanasiou, ``{OSMRec
  tool for automatic recommendation of categories on spatial entities in
  OpenStreetMap},'' in \emph{Proceedings of the 9th ACM Conference on
  Recommender Systems}.\hskip 1em plus 0.5em minus 0.4em\relax ACM, 2015, pp.
  337--338.

\bibitem{Vandecasteele_2015improving}
A.~Vandecasteele and R.~Devillers, ``{Improving volunteered geographic
  information quality using a tag recommender system: the case of
  OpenStreetMap},'' in \emph{OpenStreetMap in GIScience}.\hskip 1em plus 0.5em
  minus 0.4em\relax Springer, 2015, pp. 59--80.

\bibitem{Ali_2014}
A.~L. Ali, F.~Schmid, R.~Al-Salman, and T.~Kauppinen, ``Ambiguity and
  plausibility: managing classification quality in volunteered geographic
  information,'' in \emph{Proceedings of the 22nd ACM SIGSPATIAL international
  conference on advances in geographic information systems}, 2014, pp.
  143--152.

\bibitem{Egenhofer_1995}
M.~J. Egenhofer and R.~D. Franzosa, ``On the equivalence of topological
  relations,'' \emph{International Journal of Geographical Information
  Systems}, vol.~9, no.~2, pp. 133--152, 1995.

\bibitem{Ali_2017rule}
A.~L. Ali, Z.~Falomir, F.~Schmid, and C.~Freksa, ``{Rule-guided human
  classification of Volunteered Geographic Information},'' \emph{ISPRS Journal
  of Photogrammetry and Remote Sensing}, vol. 127, pp. 3--15, 2017.

\bibitem{Agrawal_1994}
R.~Agrawal and R.~Srikant, ``{Fast Algorithms for Mining Association Rules in
  Large Databases},'' in \emph{Proceedings of the 20th International Conference
  on Very Large Data Bases}, ser. VLDB '94.\hskip 1em plus 0.5em minus
  0.4em\relax Morgan Kaufmann Publishers Inc., 1994, pp. 487--499.

\bibitem{Kashian_2019}
A.~Kashian, A.~Rajabifard, K.~F. Richter, and Y.~Chen, ``{Automatic analysis of
  positional plausibility for points of interest in OpenStreetMap using
  coexistence patterns},'' \emph{International Journal of Geographical
  Information Science}, vol.~33, no.~7, pp. 1420--1443, 2019.

\bibitem{Funke_2017}
S.~Funke and S.~Storandt, ``{Automatic Tag Enrichment for Points-of-Interest in
  Open Street Map},'' in \emph{Web and Wireless Geographical Information
  Systems}.\hskip 1em plus 0.5em minus 0.4em\relax Cham: Springer International
  Publishing, 2017, pp. 3--18.

\bibitem{Assessment_2015arsanjani}
J.~J. Arsanjani and E.~Vaz, ``An assessment of a collaborative mapping approach
  for exploring land use patterns for several european metropolises,''
  \emph{International Journal of Applied Earth Observation and Geoinformation},
  vol.~35, pp. 329--337, 2015.

\bibitem{Integrating_2016johnson}
B.~A. Johnson and K.~Iizuka, ``{Integrating OpenStreetMap crowdsourced data and
  Landsat time-series imagery for rapid land use/land cover (LULC) mapping:
  Case study of the Laguna de Bay area of the Philippines},'' \emph{Applied
  Geography}, vol.~67, pp. 140--149, 2016.

\bibitem{Kaiser_2017}
P.~Kaiser, J.~D. Wegner, A.~Lucchi, M.~Jaggi, T.~Hofmann, and K.~Schindler,
  ``Learning aerial image segmentation from online maps,'' \emph{IEEE
  Transactions on Geoscience and Remote Sensing}, vol.~55, no.~11, pp.
  6054--6068, 2017.

\bibitem{Tuia_2017landusenorth}
D.~{Tuia}, G.~{Moser}, M.~{Wurm}, and H.~{Taubenb{\"{o}}ck}, ``{Land use
  modeling in North Rhine-Westphalia with interaction and scaling laws},'' in
  \emph{Joint Urban Remote Sensing Event (JURSE)}, March 2017, pp. 1--4.

\bibitem{Srivastava_2018agile}
S.~Srivastava, J.~E. Vargas-Mu{\~{n}}oz, S.~Lobry, and D.~Tuia, ``{Land-use
  characterisation using Google Street View pictures and OpenStreetMap},''
  2018.

\bibitem{Maggiori_2017cnn}
E.~Maggiori, Y.~Tarabalka, G.~Charpiat, and P.~Alliez, ``{Convolutional Neural
  Networks for Large-Scale Remote-Sensing Image Classification},'' \emph{IEEE
  Transaction on Geoscience and Remote Sensing}, vol.~55, no.~2, pp. 645--657,
  Feb 2017.

\bibitem{Yuan_2018}
J.~Yuan, P.~K.~R. Chowdhury, J.~McKee, H.~L. Yang, J.~Weaver, and B.~Bhaduri,
  ``{Exploiting deep learning and volunteered geographic information for
  mapping buildings in Kano, Nigeria},'' \emph{Scientific data}, vol.~5, p.
  180217, 2018.

\bibitem{Maggiori_2017recurrent}
E.~Maggiori, G.~Charpiat, Y.~Tarabalka, and P.~Alliez, ``Recurrent neural
  networks to correct satellite image classification maps,'' \emph{IEEE
  Transactions on Geoscience and Remote Sensing}, vol.~55, no.~9, pp.
  4962--4971, 2017.

\bibitem{Marcos_2018cvpr}
D.~Marcos, D.~{Tuia}, B.~Kellenberger, L.~Zhang, M.~Bai, R.~Liao, and
  R.~Urtasun, ``Learning deep structure active contours end-to-end,'' in
  \emph{Computer Vision and Pattern Recognition (CVPR)}, 2018.

\bibitem{Tasar_2018}
O.~Tasar, E.~Maggiori, P.~Alliez, and Y.~Tarabalka, ``{Polygonization of Binary
  Classification Maps Using Mesh Approximation with Right Angle Regularity},''
  in \emph{IEEE International Geoscience and Remote Sensing Symposium
  (IGARSS)}, 7 2018, pp. 6404--6407.

\bibitem{Maggiori_2017polygonization}
E.~Maggiori, Y.~Tarabalka, G.~Charpiat, and P.~Alliez, ``Polygonization of
  remote sensing classification maps by mesh approximation,'' in \emph{IEEE
  International Conference on Image Processing (ICIP)}, 2017, pp. 560--564.

\bibitem{Chen_2017}
J.~Chen and A.~Zipf, ``{DeepVGI: Deep learning with volunteered geographic
  information},'' in \emph{Proceedings of the 26th International Conference on
  World Wide Web Companion}.\hskip 1em plus 0.5em minus 0.4em\relax
  International World Wide Web Conferences Steering Committee, 2017, pp.
  771--772.

\bibitem{Chen_2018}
J.~Chen, Y.~Zhou, A.~Zipf, and H.~Fan, ``{Deep Learning From Multiple Crowds: A
  Case Study of Humanitarian Mapping},'' \emph{IEEE Transactions on Geoscience
  and Remote Sensing}, pp. 1--10, 2018.

\bibitem{Luxen_2011}
D.~Luxen and C.~Vetter, ``{Real-time Routing with OpenStreetMap Data},'' in
  \emph{Proceedings of the 19th ACM SIGSPATIAL International Conference on
  Advances in Geographic Information Systems}, ser. GIS '11, 2011, pp.
  513--516.

\bibitem{Graser_2016}
A.~Graser, ``{Integrating Open Spaces into OpenStreetMap Routing Graphs for
  Realistic Crossing Behaviour in Pedestrian Navigation},'' vol.~1, pp.
  217--230, 06 2016.

\bibitem{Hahmann_2018}
S.~Hahmann, J.~Miksch, B.~Resch, J.~Lauer, and A.~Zipf, ``{Routing through open
  spaces – A performance comparison of algorithms},'' \emph{Geo-spatial
  Information Science}, vol.~21, no.~3, pp. 247--256, 2018.

\bibitem{Floros_2013}
G.~Floros, B.~van~der Zander, and B.~Leibe, ``{OpenStreetSLAM: Global vehicle
  localization using OpenStreetMaps},'' in \emph{2013 IEEE International
  Conference on Robotics and Automation}, May 2013, pp. 1054--1059.

\bibitem{Suger_2017}
B.~Suger and W.~Burgard, ``{Global outer-urban navigation with
  OpenStreetMap},'' in \emph{IEEE International Conference on Robotics and
  Automation (ICRA)}, May 2017, pp. 1417--1422.

\bibitem{Xu_2016cross}
F.~F. Xu, B.~Y. Lin, Q.~Lu, Y.~Huang, and K.~Q. Zhu, ``{Cross-region traffic
  prediction for China on OpenStreetMap},'' in \emph{Proceedings of the 9th ACM
  SIGSPATIAL International Workshop on Computational Transportation
  Science}.\hskip 1em plus 0.5em minus 0.4em\relax ACM, 2016, pp. 37--42.

\bibitem{Ren_2019_deep_road}
Y.~Ren, T.~Cheng, and Y.~Zhang, ``Deep spatio-temporal residual neural networks
  for road-network-based data modeling,'' \emph{International Journal of
  Geographical Information Science}, vol.~0, no.~0, pp. 1--19, 2019.

\bibitem{He_2016deep}
K.~He, X.~Zhang, S.~Ren, and J.~Sun, ``Deep residual learning for image
  recognition,'' in \emph{IEEE Conference on Computer Vision and Pattern
  Recognition}, 2016, pp. 770--778.

\bibitem{Bakillah_2014fine}
M.~Bakillah, S.~Liang, A.~Mobasheri, J.~Jokar~Arsanjani, and A.~Zipf,
  ``{Fine-resolution population mapping using OpenStreetMap
  points-of-interest},'' \emph{International Journal of Geographical
  Information Science}, vol.~28, no.~9, pp. 1940--1963, 2014.

\bibitem{Yao_2017mapping}
Y.~Yao, X.~Liu, X.~Li, J.~Zhang, Z.~Liang, K.~Mai, and Y.~Zhang, ``Mapping
  fine-scale population distributions at the building level by integrating
  multisource geospatial big data,'' \emph{International Journal of
  Geographical Information Science}, vol.~31, no.~6, pp. 1220--1244, 2017.

\bibitem{Gervasoni_2018convolutional}
L.~Gervasoni, S.~Fenet, R.~Perrier, and P.~Sturm, ``Convolutional neural
  networks for disaggregated population mapping using open data,'' in
  \emph{2018 IEEE 5th International Conference on Data Science and Advanced
  Analytics (DSAA)}.\hskip 1em plus 0.5em minus 0.4em\relax IEEE, 2018, pp.
  594--603.

\bibitem{Touya_2017assessing}
G.~Touya, V.~Antoniou, A.~M. Olteanu-Raimond, and M.~D. Van~Damme, ``{Assessing
  crowdsourced POI quality: Combining methods based on reference data, history,
  and spatial relations},'' \emph{ISPRS International Journal of
  Geo-Information}, vol.~6, no.~3, p.~80, 2017.

\bibitem{Berriel_2017automatic}
R.~F. Berriel, F.~S. Rossi, A.~F. de~Souza, and T.~Oliveira-Santos,
  ``{Automatic large-scale data acquisition via crowdsourcing for crosswalk
  classification: A deep learning approach},'' \emph{Computers \& Graphics},
  vol.~68, pp. 32--42, 2017.

\bibitem{Ertler_2019traffic}
C.~Ertler, J.~Mislej, T.~Ollmann, L.~Porzi, and Y.~Kuang, ``Traffic sign
  detection and classification around the world,'' \emph{arXiv preprint
  arXiv:1909.04422}, 2019.

\bibitem{Haberle_2019geo}
M.~H{\"a}berle, M.~Werner, and X.~X. Zhu, ``Geo-spatial text-mining from
  twitter--a feature space analysis with a view toward building classification
  in urban regions,'' \emph{European journal of remote sensing}, vol.~52, no.
  sup2, pp. 2--11, 2019.

\bibitem{Haberle_2019building}
------, ``Building type classification from social media texts via geo-spatial
  textmining,'' in \emph{IGARSS 2019-2019 IEEE International Geoscience and
  Remote Sensing Symposium}.\hskip 1em plus 0.5em minus 0.4em\relax IEEE, 2019,
  pp. 10\,047--10\,050.

\bibitem{Bojanowski_2017enriching}
P.~Bojanowski, E.~Grave, A.~Joulin, and T.~Mikolov, ``Enriching word vectors
  with subword information,'' \emph{Transactions of the Association for
  Computational Linguistics}, vol.~5, pp. 135--146, 2017.

\bibitem{Huang_2018classification}
R.~Huang, H.~Taubenb{\"o}ck, L.~Mou, and X.~X. Zhu, ``{Classification of
  settlement types from Tweets using LDA and LSTM},'' in \emph{IGARSS 2018-2018
  IEEE International Geoscience and Remote Sensing Symposium}.\hskip 1em plus
  0.5em minus 0.4em\relax IEEE, 2018, pp. 6408--6411.

\bibitem{Workman_2017unified}
S.~Workman, M.~Zhai, D.~J. Crandall, and N.~Jacobs, ``A unified model for near
  and remote sensing,'' in \emph{Proceedings of the IEEE International
  Conference on Computer Vision}, 2017, pp. 2688--2697.

\bibitem{Zhang_2019fusion}
G.~{Zhang}, P.~{Ghamisi}, and X.~X. {Zhu}, ``Fusion of heterogeneous earth
  observation data for the classification of local climate zones,'' \emph{IEEE
  Transactions on Geoscience and Remote Sensing}, vol.~57, no.~10, pp.
  7623--7642, Oct 2019.

\bibitem{Settles_2009active}
B.~Settles, ``Active learning literature survey,'' University of
  Wisconsin-Madison Department of Computer Sciences, Tech. Rep., 2009.

\bibitem{Cra12}
M.~M. Crawford, D.~{Tuia}, and L.~H. Hyang, ``Active learning: Any value for
  classification of remotely sensed data?'' \emph{Proceedings of the IEEE},
  vol. 101, no.~3, pp. 593--608, 2013.

\bibitem{GOm11}
R.~Gomes, P.~Welinder, A.~Krause, and P.~Perona, ``Crowdclustering,'' in
  \emph{Proceedings of Advances in Neural Information Processing Systems
  ({NIPS})}, 2011.

\bibitem{Tui12c}
D.~{Tuia} and J.~Mu{\~n}oz-Mar{\'\i}, ``Learning user's confidence for active
  learning,'' \emph{{IEEE Transaction on Geoscience and Remote Sensing}},
  vol.~51, no.~2, pp. 872--880, 2013.

\bibitem{Neis_2012analyzing}
P.~Neis and A.~Zipf, ``{Analyzing the contributor activity of a volunteered
  geographic information project - The case of OpenStreetMap},'' \emph{ISPRS
  International Journal of Geo-Information}, vol.~1, no.~2, pp. 146--165, 2012.

\bibitem{Severinsen_2019vgtrust}
J.~Severinsen, M.~de~Roiste, F.~Reitsma, and E.~Hartato, ``{VGTrust: measuring
  trust for volunteered geographic information},'' \emph{International Journal
  of Geographical Information Science}, pp. 1--19, 2019.

\bibitem{Lawrence_2016nlmaps}
C.~Lawrence and S.~Riezler, ``{NLmaps: A Natural Language Interface to Query
  OpenStreetMap},'' in \emph{Proceedings of COLING 2016, the 26th International
  Conference on Computational Linguistics: System Demonstrations}, 2016, pp.
  6--10.

\bibitem{Haas_2016corpus}
C.~Haas and S.~Riezler, ``{A Corpus and Semantic Parser for Multilingual
  Natural Language Querying of OpenStreetMap},'' in \emph{Proceedings of the
  2016 Conference of the North American Chapter of the Association for
  Computational Linguistics: Human Language Technologies}, 2016, pp. 740--750.

\bibitem{Lawrence_2018improving}
C.~Lawrence and S.~Riezler, ``Improving a neural semantic parser by
  counterfactual learning from human bandit feedback,'' \emph{arXiv preprint
  arXiv:1805.01252}, 2018.

\bibitem{Andreas_2013semantic}
J.~Andreas, A.~Vlachos, and S.~Clark, ``Semantic parsing as machine
  translation,'' in \emph{Proceedings of the 51st Annual Meeting of the
  Association for Computational Linguistics (Volume 2: Short Papers)}, vol.~2,
  2013, pp. 47--52.

\bibitem{Lobry_2019}
S.~Lobry, J.~Murray, D.~Marcos, and D.~Tuia, ``{Visual Question Answering from
  Remote Sensing Images},'' in \emph{IEEE International Geoscience and Remote
  Sensing Symposium (IGARSS)}, 2019, to appear.

\bibitem{Ruta_2014semantic}
M.~Ruta, F.~Scioscia, D.~De~Filippis, S.~Ieva, M.~Binetti, and E.~Di~Sciascio,
  ``{A semantic-enhanced augmented reality tool for OpenStreetMap POI
  discovery},'' \emph{Transportation Research Procedia}, vol.~3, pp. 479--488,
  2014.

\end{thebibliography}

\begin{IEEEbiographynophoto}{John~E.~Vargas-Mu{\~{n}}oz} received the B.Sc. degree in informatics 
engineering from the National University of San Antonio Abad in Cusco, 
Cusco, Peru, in 2010, and the master's degree in computer science 
from the University of Campinas, Campinas, Brazil, in 2015.
He is currently pursuing the Ph.D. degree with the University of Campinas.
His research interests include machine learning, image processing, 
remote sensing image classification, and crowdsourced geographic information analysis.
\end{IEEEbiographynophoto}

\vspace{-10pt}
\begin{IEEEbiographynophoto}{Shivangi Srivastava} received her B.E. in Electrical Engineering at Delhi Technological University (Formerly Delhi College of Engineering), India, in 2009. In 2013, she received her European Master of Science in Nuclear Fusion and Engineering Physics and in 2015, Masters by Research in Mathematics Vision and Learning at Centrale Supl{\'e}lec, France. She commenced her Ph.D. at the University of Zurich, Switzerland in 2015 and is continuing the same at Wageningen University and Research, the Netherlands. Currently, her research focuses on applications of machine learning, in particular deep learning, along with multi-modal data (remote sensing, geospatial information, and social media) to understand urban land-use characterization. 
\end{IEEEbiographynophoto}

\vspace{-10pt}
\begin{IEEEbiographynophoto}{Devis Tuia} (S'07, M’09, SM’15) received the Ph.D in environmental sciences at the University of Lausanne, Switzerland, in 2009. He was a  Postdoc at the University of Val{\'e}ncia, the University of Colorado, Boulder, CO and EPFL Lausanne. Between 2014 and 2017, he was Assistant Professor at the University of Zurich. He is now Full Professor at the Geo-Information Science and Remote Sensing Laboratory at Wageningen University, the Netherlands. He is interested in algorithms for information extraction and data fusion of geospatial data (including remote sensing) using machine learning and computer vision. He serves as Associate Editor for IEEE TGRS and the Journal of the ISPRS.
More info on \url{http://devis.tuia.googlepages.com/}
\end{IEEEbiographynophoto}

\vspace{-10pt}
\begin{IEEEbiographynophoto}{Alexandre~X.~Falc{\~{a}}o} is full professor at the Institute of Computing, University of Campinas, Campinas, SP, Brazil. He received a B.Sc. in Electrical Engineering from the Federal University of Pernambuco, Recife, PE, Brazil, in 1988. He has worked in biomedical image processing, visualization and analysis since 1991. In 1993, he received a M.Sc. in Electrical Engineering from the University of Campinas, Campinas, SP, Brazil. During 1994-1996, he worked with the Medical Image Processing Group at the Department of Radiology, University of Pennsylvania, PA, USA, on interactive image segmentation for his doctorate. He got his doctorate in Electrical Engineering from the University of Campinas in 1996. In 1997, he worked in a project for Globo TV at a research center, CPqD-TELEBRAS in Campinas, developing methods for video quality assessment. His experience as professor of Computer Science and Engineering started in 1998 at the University of Campinas. His main research interests include image/video processing, 
visualization, and analysis; graph algorithms and dynamic programming; image annotation, organization, and retrieval; machine learning and pattern recognition; and image analysis applications in Biology, Medicine, Biometrics, Geology, and Agriculture.
\end{IEEEbiographynophoto}

\end{document}